%% file: neurips_2026.tex
\documentclass{article}

\PassOptionsToPackage{numbers}{natbib}

\usepackage[final,main]{neurips_2026}

\usepackage[utf8]{inputenc} 
\usepackage[T1]{fontenc}    
\usepackage{hyperref}       
\usepackage{url}            
\usepackage{booktabs}       
\usepackage{amsfonts}       
\usepackage{nicefrac}       
\usepackage{microtype}      
\usepackage{xcolor}         

\input{math_commands.tex}

\usepackage{graphicx}
\usepackage{amsmath}
\usepackage{amssymb}
\usepackage{amsthm}

\usepackage{algorithm}
\usepackage{algpseudocode}

\newtheorem{theorem}{Theorem}

\newtheorem{proposition}[theorem]{Proposition}

\title{Robust Amortized Simulation-Based Inference via Learned Error Models}

\author{
	Matthew O'Callaghan \\
	Institute of Astronomy (IoA), University of Cambridge \\
    Center for Astrophysics | Harvard \& Smithsonian\\
	\texttt{matthew.ocallaghan@cfa.harvard.edu}
	\And
	Kaisey S. Mandel \\
	IoA, University of Cambridge \\
	Statistical Laboratory, University of Cambridge \\
	Kavli Institute for Cosmology, University of Cambridge
	\And
	Gerry Gilmore \\
	IoA, University of Cambridge \\
	Institute of Astrophysics, FORTH
}

\hypersetup{hidelinks}

\begin{document}

	\maketitle
	\begin{abstract}
		Recent advances in neural density estimation have enabled amortized Bayesian inference for complex stochastic simulators. However, these methods rely on simulators accurately reflecting the true data-generating process and can degrade significantly under model misspecification. We consider the setting where multiple unlabeled observations are available and introduce robust variational neural posterior estimation (RVNP), an amortized Bayesian inference method that uses an importance-weighted autoencoder to jointly learn a misspecification-robust posterior proxy and an explicit error model that captures the simulation-to-reality gap. Our results show that RVNP can recover robust posterior inference in a data-driven and interpretable manner, outperforming previous methods on multiple metrics across misspecified benchmarks.
		
		
	\end{abstract}
	
	\section{Introduction}

Stochastic simulation models are widely used in the natural sciences and engineering to approximate complex real-world data-generating processes (DGPs). Often, the conditional density $p(\boldsymbol{x}\mid \boldsymbol{\theta})$ associated with a stochastic simulator is not available in closed form, rendering traditional Bayesian inference methods infeasible. Simulation-based inference (SBI) \cite{Cranmer_2020}, also known as \textit{likelihood-free inference}, aims to recover the posterior distribution $p(\boldsymbol{\theta}\mid \boldsymbol{x}_{\rm obs})$ for such intractable models. Advances in probabilistic machine learning have allowed neural SBI methods to be developed that can be amortized over many observations. Neural SBI has been adopted in fields such as astronomy \citep{PhysRevD.105.063017,o2026data}, particle physics \citep{atlascollaboration2024implementationneuralsimulationbasedinference}, cosmology \citep{Lemos_2023,zeghal2024simulationbasedinferencebenchmarklsst}, and neuroscience \citep{10.7554/eLife.54997,2024MLS&T...5c5019H}. However, recent work has shown that these methods can produce overconfident posterior approximations \citep{hermans2022a} and degrade substantially when the true data-generating distribution $p^*(\boldsymbol{x})$ does not belong to the family of distributions defined by the model, known as {model misspecification} \citep{cannon2022investigating,siahkoohi2023reliable,schmitt2024detecting,frazier2024statistical,kelly2025simulation}, fueling concern for their use in real-world applications.

	Misspecification in neural SBI is often addressed through a divergence-based discrepancy $\epsilon = \mathcal{D}\!\left(p^*(\boldsymbol{x}),\, p_{\rm sim}(\boldsymbol{x})\right)$ --- known as the simulation-to-reality gap --- between the true data-generating distribution $p^*(\boldsymbol{x})$ and the prior predictive distribution induced by the simulator, $p_{\rm sim}(\boldsymbol{x}) = \mathbb{E}_{p(\boldsymbol{\theta})}\!\left[p(\boldsymbol{x}\mid \boldsymbol{\theta})\right] $, where $p(\boldsymbol{\theta})$ is the prior distribution \citep{schmitt2024detecting,elsemuller2025doesunsuperviseddomainadaptation,mishra2025robust}. Many of the early approaches addressed misspecification in the \textit{non-amortized} case by accounting for related divergence-based discrepancies using single datasets \citep{ward2022robust,huang2023learning,kelly2024misspecificationrobust}. Estimating $\epsilon$ from a set of observations $\boldsymbol{x}_{\rm obs} = \{\boldsymbol{x}_i\}_{i=1}^n$ depends on the empirical density $\hat p^*(\boldsymbol{x}) = \frac{1}{n}\sum_{i=1}^n \delta_{\boldsymbol{x}_i}(\boldsymbol{x})$ reliably approximating the true distribution $p^*(\boldsymbol{x})$. However, in the non-amortized case, a single observation will only provide weak information about $\epsilon$, mostly relying on the inference observation appearing out-of-distribution (OOD) relative to the simulator prior predictive samples. In amortized SBI --- where we have multiple observations --- the empirical density $\hat p^*(\boldsymbol{x})$ provides a more reliable approximation to $p^*(\boldsymbol{x})$ \cite[Sec.~19.2, p.~269]{vaart1998asymptotic}, and thus a better estimation of $\epsilon$. This motivates methods that use a set of real, unlabeled observations to constrain the true DGP and recover better posterior approximations.

\begin{figure}
	\centering
	\includegraphics[width=\linewidth]{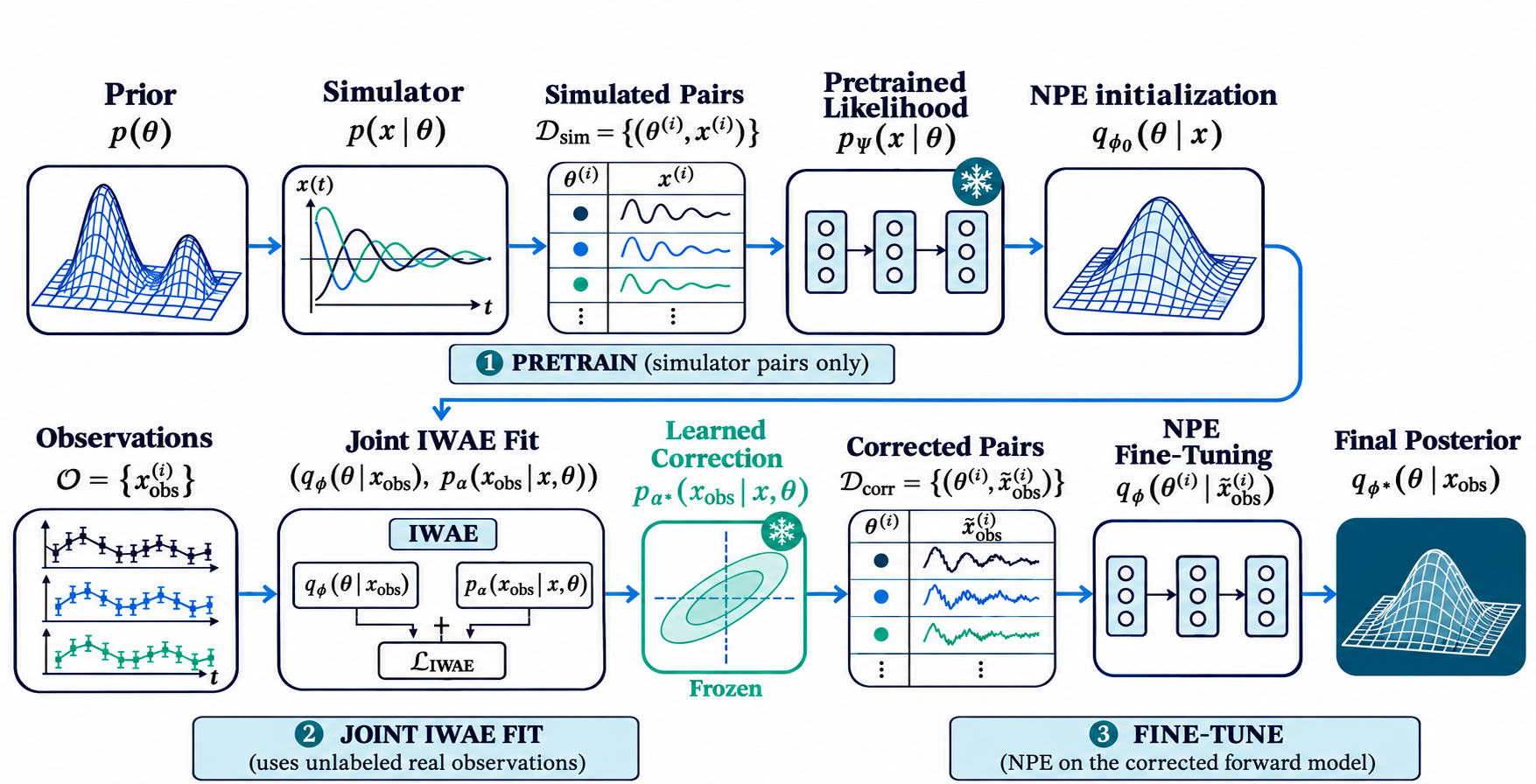}
	\caption{\textbf{RVNP training and inference workflow.} A simulator likelihood and posterior are pretrained on simulated pairs. The error model and posterior are then fitted jointly using unlabeled observations, followed by posterior fine-tuning with the learned correction fixed.}
	\label{fig:diagram}
\end{figure}
	
		Existing robust SBI approaches face challenges in misspecification-robust and interpretable \textit{amortized} inference. In particular, existing error-modeling and correction-parameter approaches \citep{frazier2021robust,ward2022robust,kelly2024misspecificationrobust} scale poorly to amortized SBI due to their reliance on Markov Chain Monte Carlo (MCMC) sampling or fixed, pre-defined error models \citep{kelly2025simulation}. Robust summary statistics using unsupervised domain adaptation (UDA) \citep{johansson2019support} and regularization techniques \citep{huang2023learning} scale more favorably to amortized SBI \citep{elsemuller2025doesunsuperviseddomainadaptation,mishra2025robust}, but they may come at the expense of a non-Bayesian interpretation of the domain-adaptation hyperparameters, limited interpretability of how the domain adaptation closes the simulation-to-reality gap \citep{elsemuller2025doesunsuperviseddomainadaptation}, and unclear trade-offs between domain adaptation and the inference algorithm \citep{chen2021neuralapproximatesufficientstatistics}. Generalized Bayesian inference methods \cite{gao2023generalized,pacchiardi2024generalized} can offer a partially amortized approach, but still rely on a sampling step and deviate from a standard likelihood-based posterior update. Fitting the simulation-to-reality gap requires an inductive bias to determine how distributional shifts between observed and simulated data are treated. This is particularly important when misspecification occurs within the support of the simulator, where model errors may still produce summary statistics that appear in-distribution (ID) relative to simulated data \citep{schmitt2024detecting,wehenkel2025addressing,elsemuller2025doesunsuperviseddomainadaptation}. 
        Recently, methods have emerged which incorporate small calibration sets from the true distribution to adapt the misspecified posterior distribution \cite{wehenkel2025addressing}. However, the existence of such a set is a strong assumption that is often not met in practice. Combining the scalability of amortized methods with the interpretability of error-model and correction-parameter approaches is an unaddressed research avenue that we investigate in this paper.

\textbf{Contributions:}
We propose robust variational neural posterior estimation (RVNP, see Figure \ref{fig:diagram} for an overview), an amortized Bayesian inference method for misspecified simulators that integrates neural posterior estimation with explicit error modeling. RVNP pre-trains a simulator likelihood, introduces a flexible error model, and optimizes an importance-weighted autoencoder (IWAE) objective \citep{burda2015importance} to maximize the evidence of observed data under the variational posterior $q_{\boldsymbol{\phi}}(\boldsymbol{\theta}\mid\boldsymbol{x}_{\rm obs})$. This formulation leverages unlabeled observations to fit the simulation-to-reality gap with an adopted error-model family, and does not rely on calibration datasets \cite{wehenkel2025addressing}, manually tuned trade-offs between posterior accuracy and domain adaptation \cite{huang2023learning,elsemuller2025doesunsuperviseddomainadaptation}, or a loss of amortization \cite{ward2022robust,huang2023learning,kelly2024misspecificationrobust}. To our knowledge, RVNP is the first amortized variational inference approach that uses error modeling to explicitly address misspecification in simulation-based inference, and the first to employ IWAE objectives in this setting. Empirically, RVNP improves posterior calibration across misspecified benchmark tasks while simultaneously refining the underlying generative forward model.

	In Section \ref{background_section}, we review the relevant SBI and misspecification background. Section \ref{methods} introduces RVNP, Section \ref{results} evaluates it empirically, and Section \ref{discussion} discusses related work, limitations, and conclusions. Furthermore, Appendix \ref{app:ablations} studies robustness versus the number of observations and additional ablations, while Appendix \ref{app:wellspecified_sir} evaluates the well-specified setting. Additional implementation details are provided in Appendix \ref{app:training_steps}.

	\section{Background}\label{background_section}
	\subsection{Simulation-based Inference}
	
	Classical approaches to SBI such
	as approximate Bayesian computation (ABC) \citep{10.1214/aos/1176346785,abc_genetics}
	and Bayesian synthetic likelihood (BSL) \citep{Price02012018} perform inference
	separately for each observation, often requiring many calls to the simulator. However, advances in neural conditional density estimation have enabled neural SBI methods that use
	simulated pairs $(\boldsymbol{\theta},\boldsymbol{x})\sim
	p(\boldsymbol{\theta})p(\boldsymbol{x}\mid\boldsymbol{\theta})$ to learn reusable,
	amortized inference networks. Neural posterior estimation (NPE) \citep{2016arXiv160506376P} trains a conditional
	density estimator $q_{\boldsymbol{\phi}}(\boldsymbol{\theta}|\,\boldsymbol{x})$ by minimizing
	\begin{equation}\label{npe_loss}
		\mathcal{L}_{\rm NPE}(\boldsymbol{\phi})
		:=
		-\mathbb{E}_{p(\boldsymbol{\theta})p(\boldsymbol{x}\mid\boldsymbol{\theta})}
		\left[
		\log q_{\boldsymbol{\phi}}(\boldsymbol{\theta}|\,\boldsymbol{x})
		\right]
	\end{equation}
	with respect to $\boldsymbol{\phi}$. Minimizing the NPE objective is equivalent, up to a constant, to minimizing the expected forward Kullback-Leibler (KL) divergence between the simulator posterior and its approximation, with the expectation taken over the simulator marginal $p(\boldsymbol{x})=\int p(\boldsymbol{x}|\, \boldsymbol{\theta})p(\boldsymbol{\theta})d\boldsymbol{\theta}$. This targets the true posterior when the simulator likelihood and prior correctly describe the data-generating process \citep{schmitt2024detecting}.
	Once trained, NPE produces an approximate posterior for a new observation in a
	single forward pass \citep{2016arXiv160506376P,NIPS2017_addfa9b7,
		greenberg2019automaticposteriortransformationlikelihoodfree,radev2022bayesflow}.
	Neural likelihood estimation (NLE) instead learns a surrogate likelihood
	$p_{\boldsymbol{\Psi}}(\boldsymbol{x}\mid\boldsymbol{\theta})$ and combines it
	with the prior using MCMC or variational inference, making it only partially
	amortized \citep{pmlr-v96-lueckmann19a,papamakarios2019sequentialneurallikelihoodfast,
		boelts2022flexible}. Related approaches include neural ratio estimation (NRE),
	which learns likelihood-to-evidence or likelihood-ratio estimators
	\citep{pmlr-v33-izbicki14,cranmer2016approximatinglikelihoodratioscalibrated,
		hermans2020likelihoodfreemcmcamortizedapproximate}.
	
	For high-dimensional observations, SBI methods often operate on summary statistics
	or learned embeddings from a neural statistic estimator (NSE) $S_{\boldsymbol{\omega}}(\boldsymbol{x})$ rather than raw
	data. These summaries can improve efficiency, but may lose information when they
	are not sufficient for $\boldsymbol{\theta}$ \citep{blum2013comparative,
		alsing2018massive}. They may be hand-crafted, pre-trained, or learned jointly with
	the inference network, and remain important in recent neural SBI methods based on
	normalizing flows, diffusion, flow matching, and consistency models
	\citep{wildberger2023flow,geffner2023compositional,sharrock2024sequential,
		gloeckler2024allinonesimulationbasedinference}. In NPE, the NSE parameters can be trained jointly but NLE requires embeddings and the likelihood to be trained separately \citep{brehmer2020flows}, although this has been shown to not be restrictive \citep{chen2021neuralapproximatesufficientstatistics,makinen2024hybrid}.

\subsection{Misspecification in SBI}
A generative model is defined as misspecified when the true data-generating process (DGP) $p^*(\boldsymbol{x})$ does not belong to the model family defined by the simulator,
$p^*(\boldsymbol{x}) \notin \mathcal{M} := \left\{p(\boldsymbol{x}\mid\boldsymbol{\theta}) : \boldsymbol{\theta} \in \Theta\right\}$,
meaning there exists no parameter value $\boldsymbol{\theta} \in \Theta$ for which $p^*(\boldsymbol{x}) = p(\boldsymbol{x}\mid\boldsymbol{\theta})$
\citep{frazier2021robust,cannon2022investigating,kelly2025simulation}.

In the likelihood-free case, simulation-based training takes the expectation with respect to the model-implied distributions, not necessarily the true data-generating distribution, and convergence does not guarantee accurate posteriors \citep{schmitt2024detecting}, particularly when the forward model is misspecified. In this paper, we address misspecification through the discrepancy between the observed data distribution and the simulator prior predictive distribution $p_{\rm sim}(\boldsymbol{x}) := \int
p(\boldsymbol{x}\mid \boldsymbol{\theta})p(\boldsymbol{\theta})\,d\boldsymbol{\theta}$
via the divergence $\epsilon_{\mathcal{D}}:=\mathcal{D}(p^*(\boldsymbol{x}),p_{\rm sim}(\boldsymbol{x}))$ \citep{mishra2025robust,elsemuller2025doesunsuperviseddomainadaptation}.
We call $\epsilon_{\mathcal{D}}$ the $\mathcal{D}$-simulation-to-reality gap.

Accounting for the marginal gap $\epsilon_{\mathcal{D}}$ alone does not automatically determine whether meaningful corrections can be made to the posterior. This limitation arises from not having any extra information about $\boldsymbol{\theta}$, and complex methods for accounting for this gap, such as unsupervised domain adaptation (UDA), can corrupt the definition of $\boldsymbol{\theta}$ in favor of posterior predictive performance \citep{elsemuller2025doesunsuperviseddomainadaptation}. In this paper, we assume that the simulator parameters are physically meaningful and that the true DGP differs from the simulator through an unmodeled corruption that retains the meaning of $\boldsymbol{\theta}$. 



\subsection{Importance-weighted Autoencoders}
Importance-weighted autoencoders (IWAE) \cite{burda2015importance} are a variant of variational autoencoders (VAEs) \citep{kingma2013auto} that use importance sampling to obtain a tighter bound on the marginal likelihood. VAEs assume a latent space $\Theta \subseteq \mathbb{R}^m$ and train a generative model $p_{\boldsymbol{\alpha}}(\bm{x}\mid \bm{\theta})$ and an approximate posterior $q_{\boldsymbol{\phi}}(\bm{\theta}\mid \bm{x})$, referred to as the decoder and encoder, by maximizing the evidence lower bound (ELBO). In this work, we optimize the IWAE objective
	\begin{align}\label{iwae}
	\log p_{\boldsymbol{\alpha}} (\boldsymbol{x} ) &\approx \log \left[ \frac {1}{k} \sum_{l=1}^{k}
	\frac {p_{\boldsymbol{\alpha}} \left(\textbf{x} , \boldsymbol{\theta}^{ (l)}
		\right)} {q_{\boldsymbol{\phi}} \left( \boldsymbol{\theta}^{(l)} | \textbf{x}
		\right)}\right] = {\mathcal{L}}^{\text{IWAE}}_k \left( \boldsymbol{\alpha},
	\boldsymbol{\phi}; \textbf{x} \right), \quad \boldsymbol{\theta}^{(l)} \sim q_{\boldsymbol{\phi}} \left( \boldsymbol{\theta} |
	\textbf{x} \right),
\end{align}
with respect to $\boldsymbol{\alpha}$ and $\boldsymbol{\phi}$, where $p(\boldsymbol{\theta})$ is a prior over the latent space. This objective is strictly tighter than the ELBO and encourages mass-covering behavior, but suffers from a low signal-to-noise ratio in gradient estimates \citep{2018arXiv180204537R}. To mitigate this, we adopt the partially importance-weighted autoencoder (PIWAE) estimator \citep{2018arXiv180204537R}, which yields more stable optimization and improved convergence. Details are provided in Appendix \ref{app:piwae}.
	

	\section{Method}\label{methods}
	We propose robust variational neural posterior estimation (RVNP), a misspecification-robust, amortized SBI method that is designed to fit the simulation-to-reality gap
	$
		\epsilon_\mathcal{D}=\mathcal{D}(p^*(\boldsymbol{x}),p_{\rm sim}(\boldsymbol{x}))
$
	when $\mathcal{D}$ is the KL divergence, by relating the observed distribution $p^*$ to simulator outputs through an auxiliary variable $\boldsymbol{x}_{\rm obs}$ and an error model $p_{\boldsymbol{\alpha}}(\boldsymbol{x}_{\rm obs}\mid\boldsymbol{x},\boldsymbol{\theta})$. The resulting amortized posterior $q_{\boldsymbol{\phi}}(\boldsymbol{\theta}\mid \boldsymbol{x}_{\rm obs})$ retains $\boldsymbol{\theta}$ interpretability and is robust when the fitted error model can explain the simulator-to-reality gap through providing a better understanding of the forward model. RVNP training first pretrains a simulator likelihood surrogate and warm-starts the posterior with NPE on simulated data, then jointly optimizes the RVNP variational objective before fine-tuning the posterior under the fixed error-adjusted forward model. An overview of the training can be found in Algorithm \ref{alg:rvnp_training}.



	\subsection{Generative model}
	 RVNP first trains a conditional density estimator $p_{\boldsymbol{\Psi}}(\boldsymbol{x}\mid \boldsymbol{\theta})$ to approximate the simulator likelihood via NLE. After fixing the parameters $\boldsymbol{\Psi}$, we model the observed DGP as:
	\begin{equation}
	(\boldsymbol{\theta},\boldsymbol{x},\boldsymbol{x}_{\rm obs})\sim p(\boldsymbol{\theta})p_{\boldsymbol{\Psi}}(\boldsymbol{x}|\,\boldsymbol{\theta})p_{\boldsymbol{\alpha}}(\boldsymbol{x}_{\rm obs}|\,\boldsymbol{x},\boldsymbol{\theta}),
	\end{equation}
	 where $p(\boldsymbol{\theta})$ is a known and tractable prior. Let $\mathcal{O}=\{\boldsymbol{x}_{\rm obs}^{(i)}\}_{i=1}^{N_{\rm obs}}$ be an unlabeled set of observations from the true DGP that we have access to during training. We assume that the observations $\boldsymbol{x}_{\rm obs}^{(i)}$ are conditionally independent given their latent parameters $\boldsymbol{\theta}^{(i)}$ and the shared error-model parameters $\boldsymbol{\alpha}$. This defines a hierarchical model in which the discrepancy between simulator outputs and observed data is parameterized through both $\boldsymbol{\theta}$ and the error model. The marginal likelihood under this model conditional on the error model parameters is given as
	\begin{equation}\label{evidence}
		p(\mathcal{O}\mid \boldsymbol{\alpha})= \prod_{i=1}^{N_{\rm obs}}\mathbb{E}_{p(\boldsymbol{\theta}^{(i)})}\mathbb{E}_{p_{\boldsymbol{\Psi}}(\boldsymbol{x}^{(i)}\mid \boldsymbol{\theta}^{(i)})}[p_{\boldsymbol{\alpha}}(\boldsymbol{x}^{(i)}_{\rm obs}\mid\boldsymbol{x}^{(i)},\bm \theta ^{(i)})].
	\end{equation}

	\subsection{RVNP Loss}
	RVNP uses an IWAE architecture to jointly learn an amortized posterior over simulator parameters together with the error model. We adopt the variational posterior $q_{\boldsymbol{\phi}}(\boldsymbol{\theta} \mid \mathbf{x}_{\rm obs})$ as the importance distribution. The resulting training objective is the empirical IWAE lower bound on the log-evidence (Equation \ref{evidence}) across the observed dataset:
	\begin{equation}\label{var_loss}
		\mathcal{L}(\boldsymbol{\phi}, \boldsymbol{\alpha})_{V}
		:=
		- \sum_{i=1}^{N_{\rm obs}}
		\log \left[
		\frac{1}{K} \sum_{l=1}^{K}
		\frac{
			\mathbb{E}_{p_{\boldsymbol{\Psi}}(\mathbf{x}^{(i)}\mid \boldsymbol{\theta}^{(l)})}
			\!\left[\, p_{\boldsymbol{\alpha}}(\boldsymbol{x}^{(i)}_{\rm obs}|\,\boldsymbol{x}^{(i)},\bm \theta ^{(l)})  \right]
			\, p(\boldsymbol{\theta}^{(l)})}
		{q_{\boldsymbol{\phi}}(\boldsymbol{\theta}^{(l)} \mid \mathbf{x}_{\rm obs}^{(i)})}
		\right]
		\end{equation}
	which we minimize with respect to $\boldsymbol{\phi}$ and $\boldsymbol{\alpha}$, where $\boldsymbol{\theta}^{(l)} \sim q_{\boldsymbol{\phi}}(\boldsymbol{\theta} \mid \mathbf{x}_{\rm obs}^{(i)})$. For each $\boldsymbol{\theta}^{(l)}$, we approximate
	$\mathbb{E}_{p_{\boldsymbol{\Psi}}(\mathbf{x}\mid \boldsymbol{\theta}^{(l)})}
	\!\left[\, p_{\boldsymbol{\alpha}}\!\left(\mathbf{x}_{\rm obs}^{(i)} \mid \mathbf{x},\boldsymbol{\theta}^{(l)}\right) \right]$ using a Monte Carlo estimate from draws from the pre-trained likelihood surrogate. Optimizing this objective requires gradients with respect to the proposal parameters $\boldsymbol{\phi}$ through samples from $q_{\boldsymbol{\phi}}(\boldsymbol{\theta}\mid\boldsymbol{x}_{\rm obs})$; in practice we parameterize $q_{\boldsymbol{\phi}}$ with normalizing flows \cite{papamakarios2021normalizingflowsprobabilisticmodeling}. A regularization term of the form $-\lambda \log p(\boldsymbol{\alpha})$ could also be added, but this introduces an explicit trade-off between fitting the simulation-to-reality gap and regularizing the error model. We omit this term in the main paper and investigate it in the appendix. 

	\subsection{Error Models}
	Error modeling and adjustment parameters are one of the earliest successful approaches to robust SBI \citep{frazier2021robust,ward2022robust,kelly2024misspecificationrobust,kelly2025simulation}. Although any forward model can be introduced within RVNP, we choose two different examples in this paper. The first error model that we adopt in RVNP includes a $\boldsymbol{\theta}$-dependent covariance matrix:
	\begin{equation}\label{rvnp_model}
		p_{\boldsymbol{\alpha}}(\boldsymbol{x}_{\rm obs}|\,\boldsymbol{x},\boldsymbol{\theta}) = \mathcal{N}(\boldsymbol{x}_{\rm obs};\,\boldsymbol{x} ,\boldsymbol{\Sigma}(\boldsymbol{\theta}\,;\,\boldsymbol{\alpha})),
	\end{equation}
	where $\boldsymbol{\Sigma}(\boldsymbol{\theta}\,;\,\boldsymbol{\alpha})$ is parameterized by a neural network that depends on $\boldsymbol{\theta}$. This allows the error model to learn local parameter-dependent corrections to the scale of the simulator outputs, thereby increasing the variance of the simulator forward model (RVNP-L). The second model we include is a global model (RVNP-G):
	\begin{equation}\label{rvnp_model_global}
		p_{\boldsymbol{\alpha}}(\boldsymbol{x}_{\rm obs}|\,\boldsymbol{x}) = \mathcal{N}(\boldsymbol{x}_{\rm obs};\,\boldsymbol{x}, \boldsymbol{\Sigma}),
	\end{equation}
	defined through a full-rank Gaussian covariance matrix $\boldsymbol{\Sigma}$ parameterized in terms of a Cholesky decomposition. This error model is constant across the parameter space and does not depend on $\boldsymbol{\theta}$. Our method generalizes very easily to include any error model that we believe explains the simulation-to-reality gap. Proposition \ref{prop:kl_projection} shows that, when $\mathcal{D}=\mathrm{KL}$, fitting the error-augmented model by maximizing log-score corresponds to a KL-minimizing fit of $p^*$ within the adopted error-model family.

	\subsection{Fine Tuning}
	Under the assumption that the learned error model provides a useful approximation to the simulation-to-reality gap, after optimizing the RVNP loss function we tune the variational posterior using an error-adjusted NPE loss \citep{Cranmer_2020, ward2022robust} 
	\begin{equation}
		\mathcal{L}(\boldsymbol{\phi})_{\text{ENPE}}
		:=-\mathbb{E}_{p(\boldsymbol{ x},\boldsymbol{ \theta})}[\mathbb{E}_{p_{\boldsymbol{ \alpha}^*}(\boldsymbol{ x}_{\text{obs}}\mid \boldsymbol{ x},\boldsymbol{ \theta})}[\log q_{\boldsymbol{\phi}}(\boldsymbol{ \theta}\mid \boldsymbol{ x}_{\rm obs})]],
	\end{equation}
	where $ \boldsymbol{\alpha}^*$ denotes the error-model parameters frozen after optimizing the IWAE objective and the first expectation is with respect to the simulator sampling distribution. 
	
	\subsection{Addressing the KL simulation-to-reality gap}
	We formalize the equivalence between the RVNP objective and fitting the KL-simulation-to-reality gap within the adopted error-model family. First define the error-augmented prior predictive distribution $
		p_{\text{model}}(\boldsymbol{x}_{\rm obs} \mid \boldsymbol{\alpha})
		=
		\mathbb{E}_{p(\boldsymbol{\theta})}
		\mathbb{E}_{p_{\boldsymbol{\Psi}}(\boldsymbol{x}\mid\boldsymbol{\theta})}
		\!\left[p_{\boldsymbol{\alpha}}(\boldsymbol{x}_{\rm obs}\mid\boldsymbol{x},\boldsymbol{\theta})\right].
$
	
	\begin{proposition}[]\label{prop:kl_projection}
		Let $p^*(\boldsymbol{x}_{\rm obs})$ denote the (unknown) data-generating distribution over observations, and assume $p_{\text{model}}(\boldsymbol{x}_{\rm obs}\mid\boldsymbol{\alpha})$ is a valid density for all $\boldsymbol{\alpha}$. Then
		\begin{equation}
			\arg\max_{\boldsymbol{\alpha}}\;
			\mathbb{E}_{p^*(\boldsymbol{x}_{\rm obs})}
			\!\left[\log p_{\text{model}}(\boldsymbol{x}_{\rm obs}\mid\boldsymbol{\alpha})\right]
			=
			\arg\min_{\boldsymbol{\alpha}}\;
			\mathrm{KL}\!\left(p^*(\boldsymbol{x}_{\rm obs})\;\|\;p_{\text{model}}(\boldsymbol{x}_{\rm obs}\mid\boldsymbol{\alpha})\right).
		\end{equation}
		Moreover, for an i.i.d.\ unlabeled observation set $\mathcal O=\{\boldsymbol{x}^{(i)}_{\rm obs}\}_{i=1}^{N_{\rm obs}}$, the empirical average log-score converges to the population log-score as $N_{\rm obs}\to\infty$ under standard regularity conditions.
	\end{proposition}
	\begin{proof}
		By definition,
		\begin{equation}
			\mathrm{KL}\!\left(p^*(\boldsymbol{x}_{\rm obs})\;\|\;p_{\text{model}}(\boldsymbol{x}_{\rm obs}\mid\boldsymbol{\alpha})\right)
			=
			\mathbb{E}_{p^*(\boldsymbol{x}_{\rm obs})}
			\!\left[\log p^*(\boldsymbol{x}_{\rm obs})\right]
			-
			\mathbb{E}_{p^*(\boldsymbol{x}_{\rm obs})}
			\!\left[\log p_{\text{model}}(\boldsymbol{x}_{\rm obs}\mid\boldsymbol{\alpha})\right].
		\end{equation}
		The first term is constant in $\boldsymbol{\alpha}$, so minimizing the KL divergence is equivalent to maximizing the population log-score. The convergence claim follows from the law of large numbers under standard regularity conditions.
	\end{proof}
 
	This standard KL identity identifies the best fit within the chosen error-augmented family, and it does not guarantee posterior coverage. RVNP therefore fits the simulation-to-reality gap by targeting the best error-augmented simulator in the chosen family, with posterior robustness following when that family can explain the corruption. RVNP assumes that $\boldsymbol{\theta}$ are physically interpretable under the model definition and the family of error models reflects the prior beliefs about how misspecification should be handled, allowing for a controllable and interpretable approach to robust inference in line with the posterior predictive approach \cite{schmitt2024detecting,mishra2025robust,elsemuller2025doesunsuperviseddomainadaptation}.

	\section{Experiments} \label{results}

	\begin{figure}[t]
		\centering
		\includegraphics[width=\textwidth,keepaspectratio]{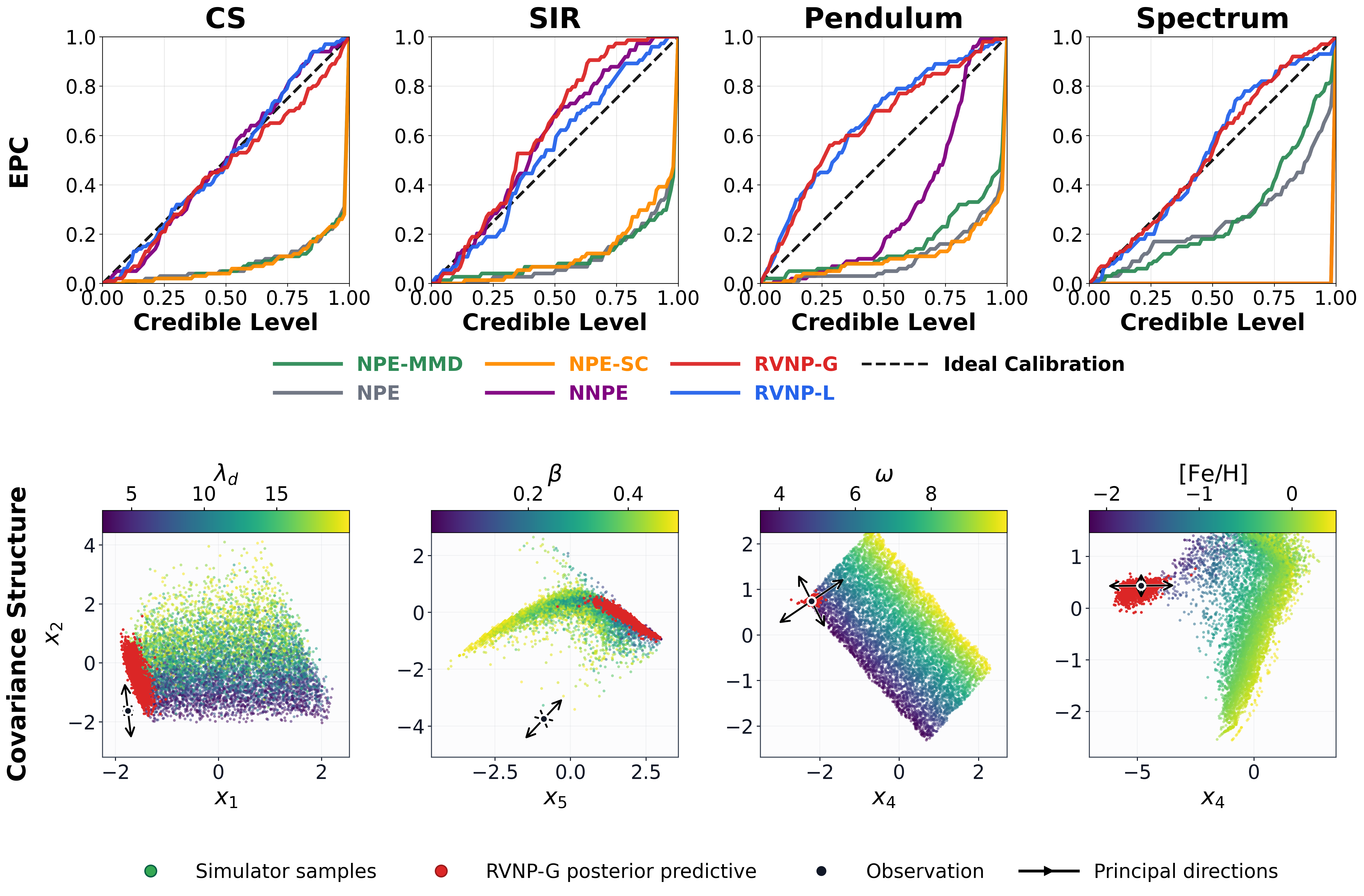}
		\caption{\textbf{Graphical diagnostic results across each experiment.} Top row: expected posterior coverage plot at different credible levels for each task across each benchmark algorithm. The diagonal indicates well-calibrated posteriors, under the diagonal indicates overconfident and above the diagonal represents conservative posteriors. Bottom row: qualitative covariance-structure diagnostics comparing simulator samples (green) with RVNP-G posterior-predictive samples (red) and indicating principal directions of the learned global error covariance, projected to two summary dimensions.}
		\label{fig:results}
	\end{figure}
\begin{table*}[t]
	\centering
	\small
    \caption{\textbf{Summary metrics across tasks.} NPE-SC denotes neural posterior estimation with a self-consistency loss \citep{mishra2025robust}. AEPC is best when closest to $0$; LPD is best when larger; PPD is best when smaller; SRMSE is best when smaller. For each task and metric (excluding PC), the best value is bolded. Posterior contraction (PC) should be interpreted jointly with calibration and accuracy rather than in isolation. Values are reported as mean $\pm$ standard deviation.}
	\setlength{\tabcolsep}{5pt}
	\begin{tabular}{llccccc}
		\hline
		Task & Method & AEPC & LPD & PPD & SRMSE & PC \\
		\hline
		CS & RVNP-L & \textbf{-0.00} & \boldmath{$-1.53 \pm 0.80$} & $1.72 \pm 0.46$ & $1.76 \pm 0.33$ & 0.48 \\
		& RVNP-G & -0.04 & $-1.74 \pm 1.16$ & $1.71 \pm 0.49$ & \boldmath{$1.75 \pm 0.31$} & 0.48 \\
		& NPE & -0.44 & $-8.98 \pm 9.37$ & $1.56 \pm 0.47$ & $1.83 \pm 0.60$ & 0.74 \\
		& NPE-MMD & -0.44 & $-8.07 \pm 7.56$ & \boldmath{$1.53   \pm 0.43$} & $1.84 \pm 0.59$ & 0.73 \\
		& NPE-SC & -0.45 & $-10.77 \pm 11.93$ & $1.64 \pm 0.49$ & $1.86 \pm 0.61$ & 0.77 \\
		& NNPE & \textbf{0.00} & $-2.31 \pm 1.25$ & $1.79 \pm 0.48$ & $1.88 \pm 0.32$ & 0.39 \\
		\hline
		SIR & RVNP-L & \textbf{0.05} & \boldmath{$1.33 \pm 1.38$} & $2.66 \pm 1.06$ & \boldmath{$0.44 \pm 0.28$} & 0.91 \\
		& RVNP-G & 0.13 & $0.47 \pm 0.80$ & $2.61 \pm 1.06$ & $0.64 \pm 0.24$ & 0.84 \\
		& NPE & -0.45 & $-6.03 \pm 11.11$ & $2.62 \pm 1.15$ & $0.67 \pm 0.67$ & 0.93 \\
		& NPE-MMD & -0.48 & $-34.57 \pm 72.80$ & {$2.60 \pm 1.11$} & $0.71 \pm 0.65$ & 0.96 \\
		& NPE-SC & -0.45 & $-10^{3} \pm 10^{4}$ & \boldmath{$2.44 \pm 1.24$} & $1.13 \pm 0.97$ & 0.99 \\
		& NNPE & 0.09 & $0.14 \pm 0.75$ & $2.65 \pm 1.08$ & $0.86 \pm 0.41$ & 0.75 \\
		\hline
		Pendulum & RVNP-L & 0.19 & $3.77 \pm 0.88$ & $0.17 \pm 0.05$ & $0.09 \pm 0.03$ & 1.00 \\
		& RVNP-G & \textbf{0.16} & \boldmath{$3.83 \pm 0.83$} & $0.17 \pm 0.04$ & $0.10 \pm 0.03$ & 1.00 \\
		& NPE & -0.45 & $0.46 \pm 6.84$ & \boldmath{$0.13 \pm 0.05$} & $0.06 \pm 0.04$ & 1.00 \\
		& NPE-MMD & -0.39 & $-0.75 \pm 6.97$ & $0.20  \pm  0.07$ & $0.11 \pm 0.06$ & 1.00 \\
		& NPE-SC & -0.41 & $-1.27 \pm 10.38$ & \boldmath{$0.13 \pm 0.05$} & \boldmath{$0.05 \pm 0.04$} & 1.00 \\
		& NNPE & -0.21 & $-1.18 \pm 1.00$ & $1.73 \pm 0.64$ & $1.21 \pm 0.40$ & 0.65 \\
		\hline
		Spectrum & RVNP-L & \textbf{0.01} & $0.21 \pm 1.97$ & $0.42 \pm 0.12$ & $1.59 \pm 0.51$ & 0.55 \\
		& RVNP-G & 0.02 & \boldmath{$1.73 \pm 1.06$} & \boldmath{$0.31 \pm 0.08$} & \boldmath{$1.45 \pm 0.53$} & 0.75 \\
		& NPE & -0.29 & $-8.15 \pm 46.38$ & $0.32 \pm 0.09$ & $1.58 \pm 0.50$ & 0.59 \\
		& NPE-MMD & -0.24 & $-1.12 \pm 5.34$ & $1.63 \pm 1.88$ & $2.63 \pm 1.21$ & 0.18 \\
		& NPE-SC & -0.49 & $-33.67 \pm 75.03$ & $6.59 \pm 0.87$ & $5.84 \pm 0.29$ & 0.98 \\
		& NNPE & -0.49 & $-22.24 \pm 38.83$ & $5.71 \pm 1.09$ & $4.70 \pm 0.26$ & 0.34 \\
		\hline
	\end{tabular}
	
	\label{tab:config-matrix-metrics}
\end{table*}
	
	We evaluate our main claim: that RVNP enables robust, interpretable amortized posterior inference under various forms of model misspecification. The experiments cover extreme OOD summary statistics, literature benchmarks, and data with weaker-than-expected signal. We also evaluate RVNP on a prior misspecification problem. 
	
	\textbf{RVNP Variants.} We test two RVNP variants: RVNP-L, the local model that uses the $\boldsymbol{\theta}$-dependent error model (Equation \ref{rvnp_model}), while RVNP-G uses the global error model (Equation \ref{rvnp_model_global}). RVNP-G and RVNP-L share the same simulator-likelihood surrogate $p_{\boldsymbol{\Psi}}$, posterior architecture $q_{\boldsymbol{\phi}}$, optimization settings, and training budgets; they differ only in the parameterization of the error model $p_{\boldsymbol{\alpha}}$. 
    
	\textbf{Benchmark Algorithms.} We benchmark RVNP against noisy neural posterior estimation (NNPE) \cite{ward2022robust}, self-consistency loss neural posterior estimation (NPE-SC) \cite{mishra2025robust}, and the neural posterior estimation with maximum mean discrepancy (NPE-MMD) domain adaptation approach \cite{elsemuller2025doesunsuperviseddomainadaptation}, together with standard NPE. These all produce inference networks that can be applied to new observations without observation-specific posterior optimization or additional sequential simulation, making them suitable for comparison to RVNP. For NPE-MMD we dynamically learn the summary statistics jointly as the misspecification robustness arises from this approach. For the other models, we use the pre-computed summary statistics or pre-trained neural summary statistics to obtain a competitive and faithful comparison. In NPE-SC we use the same pre-trained simulator likelihood as RVNP for the self-consistency loss. In each of the benchmark algorithms we use the recommended hyperparameter choices and the exact same flow architecture as RVNP. Further details about implementation can be found in Appendix \ref{app:benchmark}.

    We do not benchmark observation-specific alternatives: RNPE uses MCMC to infer a correction for each observation \citep{ward2022robust}; generalized Bayesian methods require posterior sampling even when their cost model is amortized \citep{gao2023generalized,pacchiardi2024generalized}; and robust sequential NLE uses additional simulation rounds \citep{kelly2024misspecificationrobust}. For the supplementary Spectrum task, differentiable coverage calibration and balanced NPE provide complementary approaches to posterior overconfidence \citep{falkiewicz2023calibrating,delaunoy2023balancing} but we do not evaluate them here.
    
	\textbf{Training Setup.} We assume a fixed simulation budget of $N_{\rm sim}=100{,}000$ used alongside $N_{\rm obs}=10{,}000$ real observations, of which $10\%$ is retained for validation. RVNP, NPE-SC, and NPE-MMD use both budgets fully, while NPE and NNPE use only the simulation budget. All methods are evaluated on 1000 unseen observations. In the Appendices, we report performance as a function of $N_{\rm obs} \in \{10,10^2,10^3,10^4,10^5\}$ for a single experiment, assess RVNP on a well-specified task, and study the effect of regularizing the error-model parameters. Runtime is provided in Appendix \ref{app:training_time}. Training and implementation details are provided in Appendix \ref{app:training_steps}.

	
	\subsection{Evaluation Metrics}
	Let $T=\{(\boldsymbol{\theta}^{*(i)},\boldsymbol{x}_{\rm obs}^{(i)})\}_{i=1}^{N_{\rm test}}$ denote a labeled test set from the true DGP. We evaluate: (i) the {log posterior density} ({LPD}), given by $\log q_{\boldsymbol{\phi}}(\boldsymbol{\theta}^{*(i)}\mid \boldsymbol{x}_{\rm obs}^{(i)})$, which measures posterior accuracy at the true parameter, and report its mean and standard deviation across $T$ 
	(ii) the {average expected posterior coverage} ({AEPC}), which summarizes calibration of the posterior coverage curve, where the expected posterior coverage at credible mass $\gamma$ is the probability that the true parameter lies inside the corresponding highest-density region of posterior mass $\gamma$; values near $0$ indicate good calibration, negative values indicate overconfidence, and positive values indicate underconfidence; (iii) the {standardized root mean squared error} ({SRMSE}), which measures parameter-estimation error in standardized parameter space; (iv) the {posterior predictive distance} ({PPD}) using the Euclidean metric between the observed point and the simulator predictive samples; (v) {posterior contraction} ({PC}), which measures the reduction in posterior spread relative to the prior. Complete definitions are given in Appendix \ref{app:metrics}.

	\subsection{Task Descriptions}
	We consider four tasks covering structural misspecification, observational misspecification, and prior misspecification; full simulator and implementation details are given in Appendix \ref{app:tasks} and details of pre-trained neural statistical embeddings (NSE) can be found in Appendix \ref{app:nse}.

		\textbf{Task A: CS.}
		We consider the cancer-stromal cell development model of \citet{ward2022robust}, parameterized by Poisson rates $\boldsymbol{\theta}=(\lambda_c,\lambda_p,\lambda_d)$ controlling the number of cells, parent points, and daughters per parent. We use independent uniform priors $\lambda_c\sim U(200,1500)$, $\lambda_p\sim U(3,20)$, and $\lambda_d\sim U(10,20)$. The simulator generates a 2D spatial point pattern in which cancer labels are assigned according to proximity to latent parent locations. We use $\boldsymbol{x}\in\mathbb{R}^{4}$ summary statistics representing cancer/stromal counts and nearest-neighbor distance features.\\ \textit{Misspecification:} Observed data are generated by removing cells in tumor-core regions to mimic necrosis.

		\textbf{Task B: SIR.}
		We consider the stochastic SIR epidemic model of \citet{ward2022robust}, with parameters $\boldsymbol{\theta}=(\beta,\gamma)$ denoting infection and recovery rates. We draw $\beta$ and $\gamma$ independently from $U(0,0.5)$ and reject samples with $\gamma>\beta$, giving prior support $0<\gamma\leq\beta<0.5$. The simulator generates infection trajectories with time-varying transmission, and we summarize each trajectory with $\boldsymbol{x}\in\mathbb{R}^{6}$ summary statistics capturing outbreak size, peak timing, and autocorrelation.\\ \textit{Misspecification:} Weekend reporting delays reduce recorded infections by 5\% on weekends, with the deficit reassigned to Mondays.

		\textbf{Task C: Pendulum.}
		We consider a stochastic pendulum simulator with parameters $\boldsymbol{\theta}=(\omega_0,A)$ corresponding to the fundamental frequency and amplitude. We use $\omega_0\sim U(0.5,3.0)$ and $A\sim U(3.5,10.0)$, with phase $\phi\sim U(-0.03,0.03)$ in data generation. The observation is a noisy horizontal-position time series at 200 time steps over $10\,\mathrm{s}$, $\boldsymbol{x}\in\mathbb{R}^{200}$, with a $5$\% Gaussian observation error, and we use a pre-trained NSE.\\ \textit{Misspecification:} The observed data are generated at a lower signal-to-noise ratio than assumed by the simulator, constituting a noisier observation than modeled by the simulator with a $20$\% Gaussian observation error.
	
		\textbf{Task D: Spectrum.}
		We consider a stellar spectroscopy task with parameters $\boldsymbol{\theta}=(T_{\rm eff}, \log g, \mathrm{[Fe/H]})$, where synthetic spectra are generated by combining stellar evolution and atmosphere models and a Galactic prior \cite{high_gal_me} is used for $\boldsymbol{\theta}$. The observation is a stellar spectrum $\boldsymbol{x}\in\mathbb{R}^{250}$, and we use a pre-trained NSE. \\
			\textit{Misspecification:} We generate observations from a metal-poor stellar stream \citep{martin2022stellar}. The stream's parameter distribution differs from the Galactic prior used to generate training simulations, constituting prior misspecification while the forward model remains unchanged. This creates an accompanying finite-simulation coverage gap between simulated training data and observed spectra. Although prior misspecification is not the main target of RVNP, this supplementary experiment shows that RVNP is robust under this prior shift. 
\subsection{Description of Results}

Table~\ref{tab:config-matrix-metrics} summarizes performance across the four experiments, with Figure~\ref{fig:results} visualizing some of the results. The main finding is that explicitly modeling the simulation-to-reality gap yields more robust posterior inference. RVNP-L and RVNP-G remain well calibrated, with AEPC values close to zero, while maintaining strong posterior accuracy in terms of LPD and SRMSE, together with meaningful posterior contraction. We recommend RVNP-L as the default model, since it can capture local structure while reducing to an approximately constant correction when supported by the data.

Posterior predictive checks, together with inspection of the learned error model, provide a useful diagnostic for whether RVNP is attributing discrepancy to the correction model rather than to $\boldsymbol{\theta}$. The second row of Figure~\ref{fig:results} shows examples of the principal covariance directions in the summary statistics. Additional coverage plots are given in Appendix Figure~\ref{fig:coverage}.
	\begin{figure}
		\centering
		\includegraphics[width=\linewidth]{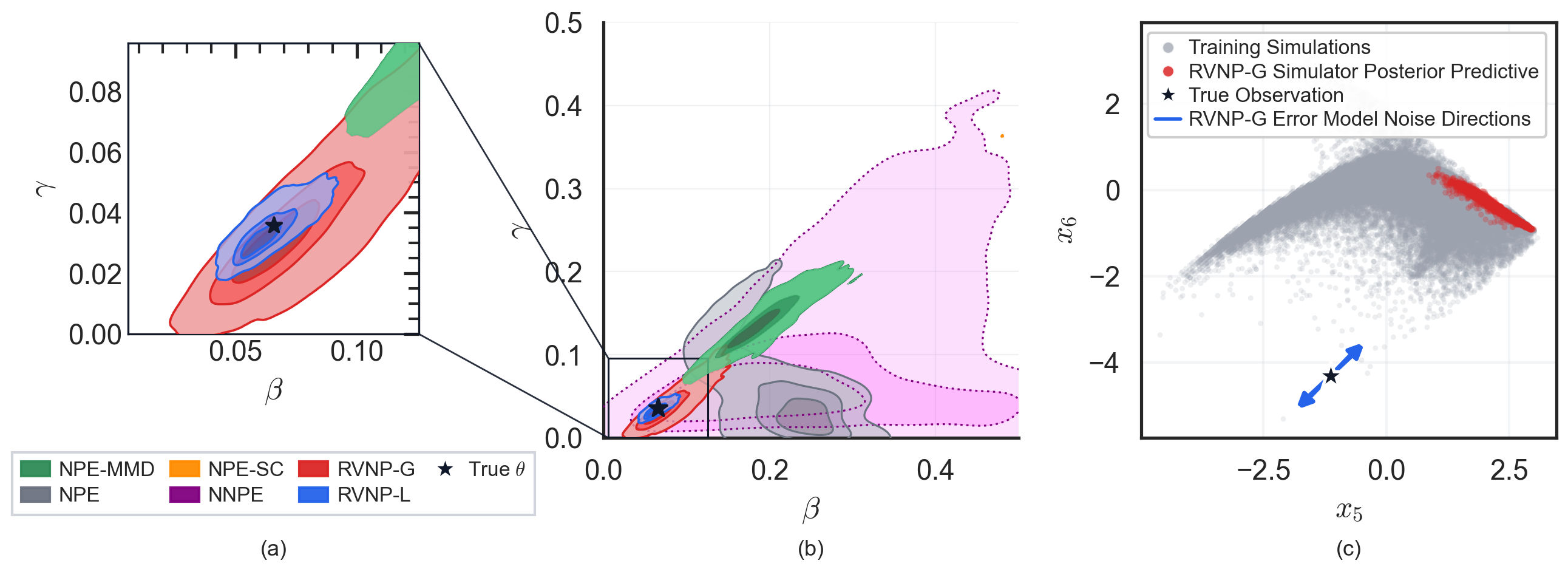}
		\caption{\textbf{Misspecified SIR example: posterior calibration and error-model interpretation.} Panels (a) and (b) compare approximate posteriors over $\boldsymbol{\theta}=(\beta,\gamma)$ for a single misspecified observation (star denotes the ground-truth parameter). RVNP variants concentrate near the truth while maintaining appropriate uncertainty in this benchmark, whereas neural posterior estimation with maximum mean discrepancy (NPE-MMD) \citep{elsemuller2025doesunsuperviseddomainadaptation} can concentrate near the truth but remain miscalibrated. Panel (c) visualizes the learned global error-model covariance (principal noise directions, blue) in summary space, alongside simulator samples (grey) and the RVNP-G posterior predictive (red), illustrating which directions the error model down-weights to fit the simulation-to-reality gap.}
		\label{fig:plot_sir_example}
	\end{figure}
    
\textbf{RVNP provides the most misspecification-robust inference overall.}
RVNP is the only method that remains close to calibrated while achieving strong posterior accuracy across all four tasks. Standard NPE, NPE-SC, and NPE-MMD can be competitive on some point or predictive metrics, but their strongly negative AEPC values indicate systematic overconfidence. NNPE is also well calibrated when its fixed corruption model matches the misspecification, but its performance degrades when the discrepancy has a different structure. Only the RVNP methods among those tested here can account for the prior misspecification in the Spectrum task. Figure~\ref{fig:plot_sir_example}(a,b) illustrates this behavior for a single observation: robust methods improve over NPE, but only RVNP both concentrates near the true value and maintains appropriate uncertainty.

\textbf{NNPE versus RVNP: fixed versus learned error models.}
NNPE obtains robustness by imposing a pre-specified corruption model \cite{ward2022robust}. This can work well when the true discrepancy aligns with the assumed corruption, but can limit contraction or lead to overconfidence otherwise. In contrast, RVNP learns the error model from the unlabeled observation set, giving it greater flexibility and more consistent calibration and accuracy across benchmarks.

\textbf{Interpretability: modeling rather than closing the gap.}
NPE-MMD attempts to close the simulation-to-reality gap through UDA, while NPE-SC uses a likelihood-based self-consistency regularization. Both can improve predictive metrics in some settings, but may adapt the posterior in ways that are not $\boldsymbol{\theta}$-information preserving, as reflected by coverage and log-probability metrics. RVNP instead models the simulation-to-reality gap explicitly through a forward error model, preserving $\boldsymbol{\theta}$ as the simulator parameter rather than forcing it to explain non-simulator variation.

\textbf{Error modeling fits the observed data under the simulator model.}
Because RVNP represents the simulation-to-reality gap explicitly, its posterior predictive checks are more interpretable than those obtained from unsupervised posterior adaptation. In CS, SIR, and Pendulum, RVNP, NPE-MMD, and NPE-SC obtain similar PPD values, while in Spectrum, NPE-MMD and NPE-SC perform poorly. However, PPD values alone can be difficult to interpret under misspecification, especially when dominated by out-of-distribution observations. The learned RVNP error model provides additional insight into this variation. We show an example in Appendix~\ref{app:interpretability} of a learned covariance matrix.

\textbf{Ablation studies and well-specified problem.}
Appendix~\ref{app:ablations} studies the robustness of RVNP to key hyperparameters and modeling assumptions, and Appendix~\ref{app:wellspecified_sir} considers a well-specified example. RVNP is stable for $N_{\rm obs}>100$, but gives poor coverage for very small datasets, which are outside its intended use case. It remains robust even with a poorly trained surrogate likelihood, and fine-tuning gives a small but consistent improvement. Strong L1 regularization of the error model harms robustness, while weak shrinkage has little effect. The number of importance samples has limited impact on metrics, although larger values improve training stability. In the well-specified setting, RVNP performs well but is slightly underconfident as shrinking the covariance to zero is difficult.
\section{Discussion and limitations}\label{discussion}
\textbf{Summary.}
RVNP is a misspecification-robust amortized SBI method that combines an explicit error model with variational inference to learn the simulation-to-reality gap within a chosen error-model family. The main result of this paper is that modeling this gap yields robust posterior inference under misspecification while retaining the scalability of amortized inference across many observations. Unlike several existing robust amortized approaches \cite{huang2023learning,ward2022robust,kelly2024misspecificationrobust,elsemuller2025doesunsuperviseddomainadaptation,mishra2025robust}, RVNP does not introduce a robustness hyperparameter trading off adaptation against posterior recovery. Instead, the correction is learned directly from unlabeled observations, and the resulting error model remains interpretable. RVNP is most useful when misspecification is expected but its form is unknown, and will benefit from progress in detecting misspecification \cite{schmitt2024detecting}. In practice, RVNP is useful when the simulator likelihood (over data or summary statistics) can be learned easily. Across our experiments, the local error model performed strongly in both global and local misspecification regimes, making it a natural default.

\textbf{Limitations.}
RVNP requires multiple training stages: learning a likelihood surrogate, fitting the error model and amortized posterior with an importance-weighted autoencoder objective, and applying a final fine-tuning step. This makes training more expensive than standard NPE baselines, although inference remains amortized and fast after training. In well-specified settings, it can also be difficult to collapse the learned covariance fully to zero, making RVNP slightly conservative.

RVNP depends on accurate upstream modeling. Since the error model is fit between the prior predictive distribution induced by the learned likelihood surrogate and the observed data distribution, a reliable surrogate helps ensure that the correction reflects genuine simulator-observation misspecification. This was not a limiting factor in our experiments, but remains an important consideration. 

Finally, in the fully unlabeled setting, RVNP inherits a fundamental limitation of misspecification-robust inference: explaining the simulation-to-reality gap requires an inductive bias. In RVNP, this bias is expressed through a practitioner-chosen error model that preserves the interpretation of $\boldsymbol{\theta}$ as the simulator parameter. Other methods account for the gap differently, but without additional information about $\boldsymbol{\theta}$ under the true data-generating process, all are constrained by their assumptions about the discrepancy. The correction family is the central inductive bias and if it cannot capture discrepancy features relevant to distinguishing $\boldsymbol{\theta}$, residual posterior bias or miscalibration may remain. This can be seen in Appendix \ref{failure_mode}.

This paper focused on two Gaussian error-model families, local and global. Other applications could use stronger inductive biases, such as sparse, low-rank, mean offsets, or physics-informed error models. This flexibility is a key strength of RVNP: the error model can be chosen to reflect scientific knowledge about the expected corruption. A broader study of such error models, or their combination with calibration data, is a natural direction for future work.

\newpage
\begin{ack}
MOC is supported by NASA under Grant No.~80NSSC26K0021. MOC was funded by the Gianna Angelopoulos Programme for Science, Technology and Innovation (GAPSTI) through the Science and Technology Facilities Council (STFC) in the UK. MOC and KSM were supported by the European Union’s Horizon 2020 research and innovation programme under ERC Grant Agreement No. 101002652 (BayeSN; PI K. Mandel) and Marie Skłodowska-Curie Grant Agreement No. 873089 (ASTROSTAT-II). This research was supported in part by the Munich Institute for Astro, Particle and BioPhysics (MIAPbP), which is funded by the Deutsche Forschungsgemeinschaft (DFG, German Research Foundation) under Germany’s Excellence Strategy – EXC-2094–390783311. GG acknowledges support from The Leverhulme Trust, through Emeritus Fellowship 2025-007.

The authors declare no competing interests.
\end{ack}

	\bibliography{main}
	\bibliographystyle{plainnat}
	\newpage

	\appendix
	
	\section{Training Details}\label{app:training_steps}

	\subsection{RVNP Training Overview}

	\textbf{Stage 1:} Pretrain the neural likelihood emulator \(p_{\boldsymbol{\Psi}}(\boldsymbol{x}\mid\boldsymbol{\theta})\) on simulated pairs \((\boldsymbol{\theta},\boldsymbol{x})\).
	
	\textbf{Stage 2:} Warm-start the amortized posterior \(q_{\boldsymbol{\phi}}(\boldsymbol{\theta}\mid\boldsymbol{x})\) with an NPE objective on simulated data.

	\textbf{Stage 3:}
	Given observed samples \(\mathcal{O}=\{\boldsymbol{x}_{\mathrm{obs}}^{(i)}\}_{i=1}^{N_{\mathrm{obs}}}\), we optimize
	\[
	(\hat{\boldsymbol{\phi}},\hat{\boldsymbol{\alpha}})
	=
	\arg\min_{\boldsymbol{\phi},\,\boldsymbol{\alpha}}
	\mathcal{L}_{\mathrm{RVNP}}
	\!\left(
	q_{\boldsymbol{\phi}},
	p_{\boldsymbol{\Psi}},
	r_{\boldsymbol{\alpha}};
	\mathcal{O}
	\right),
	\]
	where \(r_{\boldsymbol{\alpha}}\) is parameterized by \(\boldsymbol{\Sigma}(\boldsymbol{\theta};\boldsymbol{\alpha})\).

	\textbf{Stage 4:} Freeze the learned error-model parameters $\boldsymbol{\alpha}^{\star}$.

	\textbf{Stage 5 (corrupted posterior fine-tuning):}
	Fine-tune posterior parameters with an error-adjusted NPE objective
	\citep{Cranmer_2020,ward2022robust}:
	\begin{equation}
		\mathcal{L}_{\mathrm{ENPE}}(\boldsymbol{\phi})
		=
		-\mathbb{E}_{p(\boldsymbol{x},\boldsymbol{\theta})}
		\!\left[
		\mathbb{E}_{p_{\boldsymbol{\alpha}^{\star}}(\boldsymbol{x}_{\mathrm{obs}}\mid\boldsymbol{x},\boldsymbol{\theta})}
		\!\left[
		\log q_{\boldsymbol{\phi}}(\boldsymbol{\theta}\mid\boldsymbol{x}_{\mathrm{obs}})
		\right]
		\right].
	\end{equation}

	\begin{algorithm}[t]
			\caption{\textbf{RVNP training.}}
		\label{alg:rvnp_training}
		\small
		\begin{algorithmic}[1]
			\Require \(p(\boldsymbol{\theta})\), \(p_{\boldsymbol{\Psi}}(\boldsymbol{x}\mid\boldsymbol{\theta})\)
			\Require \(\mathcal O=\{\boldsymbol{x}_{\mathrm{obs}}^{(i)}\}_{i=1}^{N_{\mathrm{obs}}}\), \(\mathcal D=\{(\boldsymbol{\theta}^{(i)},\boldsymbol{x}^{(i)})\}_{i=1}^{N_{\mathrm{train}}}\)
			\Require posterior \(q_{\boldsymbol{\phi}}(\boldsymbol{\theta}\mid\boldsymbol{x})\), correction covariance \(\boldsymbol{\Sigma}(\boldsymbol{\theta};\boldsymbol{\alpha})\)
			\State pretrain \(p_{\boldsymbol{\Psi}}\) via NLE on \(\mathcal D\)
			\State warm-start \(q_{\boldsymbol{\phi}}\) via NPE on \(\mathcal D\)
			\State \((\boldsymbol{\phi},\boldsymbol{\alpha}^{\star}) \leftarrow \textsc{RVNP\_TRAIN}(\mathcal O,T_{\mathrm{RVNP}})\)
			\State freeze the correction parameters \(\boldsymbol{\alpha}^{\star}\)

			\For{\(t=1,\dots,T_{\mathrm{ENPE}}\)}
			\State sample \(\{(\boldsymbol{\theta}_b,\boldsymbol{x}_b)\}_{b=1}^{B}\subset\mathcal D\)
			\State \(\boldsymbol{x}_{\mathrm{obs},b}\sim p_{\boldsymbol{\alpha}^{\star}}(\boldsymbol{x}_{\mathrm{obs}}\mid\boldsymbol{x},\boldsymbol{\theta})\)
			\State update \(\boldsymbol{\phi}\) via \(\mathcal L_{\mathrm{ENPE}}\)
			\EndFor
			\State $\boldsymbol{\phi}^{\star}$ $\leftarrow$ $\boldsymbol{\phi}$
			\State \Return \(\boldsymbol{\phi}^{\star},\boldsymbol{\alpha}^{\star}\)
		\end{algorithmic}
	\end{algorithm}
	
	\begin{algorithm}[t]
			\caption{\textbf{\(\textsc{RVNP\_TRAIN}\).}}
		\label{alg:rvnp_train}
		\small
		\begin{algorithmic}[1]
			\Require \(\mathcal X=\{\boldsymbol{x}_{\mathrm{obs}}^{(i)}\}_{i=1}^{N}\), iterations \(T\), \(q_{\boldsymbol{\phi}}\), \(p_{\boldsymbol{\Psi}}\), \(p_{\boldsymbol{\alpha}}\), \(K,M\)
			\For{\(t=1,\dots,T\)}
			\State sample minibatch \(\mathcal B\subset\mathcal X\)
			\ForAll{\(\boldsymbol{x}_{\mathrm{obs}}^{(i)}\in\mathcal B,\ \ell=1,\dots,K\)}
			\State \(\boldsymbol{\theta}_{i\ell}\sim q_{\boldsymbol{\phi}}(\boldsymbol{\theta}\mid\boldsymbol{x}_{\mathrm{obs}}^{(i)})\)
			\State \(\boldsymbol{x}_{i\ell m}\sim p_{\boldsymbol{\Psi}}(\boldsymbol{x}\mid\boldsymbol{\theta}_{i\ell}),\ m=1,\dots,M\)
			\EndFor
			\State evaluate \(\mathcal L_{\mathrm{RVNP}}(\boldsymbol{\phi},\boldsymbol{\alpha})\)
			\State (\(\boldsymbol{\phi},\boldsymbol{\alpha})\leftarrow \mathrm{Adam}_{\boldsymbol{\phi},\boldsymbol{\alpha}}((\boldsymbol{\phi},\boldsymbol{\alpha}),\nabla_{\boldsymbol{\phi},\boldsymbol{\alpha}}\mathcal L_{\mathrm{RVNP}})\)

			\EndFor
			\State \Return \(\boldsymbol{\phi},\boldsymbol{\alpha}\)
		\end{algorithmic}
	\end{algorithm}
	
	\subsection{Training Setup}
	In the default experiments we use $N_{\rm sim}=100,000$ simulator samples and $N_{\rm obs}=10,000$ unpaired observations.
	Unless stated otherwise, RVNP-G and RVNP-L use the same architectures, optimization hyperparameters, and training budgets throughout; the only change between the two variants is the error model (Equations \ref{rvnp_model_global} and \ref{rvnp_model}).
	For reproducibility, we fix the random seed to $42$.
	
	\paragraph{Default RVNP hyperparameters.}
	Unless otherwise stated, all RVNP experiments use \(K=4\) variational/importance samples per observation in the RVNP objective (Algorithm \ref{alg:rvnp_train}). When estimating the inner expectation over \(\boldsymbol{x}\sim p_{\boldsymbol{\Psi}}(\boldsymbol{x}\mid\boldsymbol{\theta})\), we use \(M=8\) Monte Carlo samples per \((\boldsymbol{x}_{\rm obs},\boldsymbol{\theta})\) pair.

	In the simulator density training we use $300$ epochs with a batch size of $1024$ using the Adam optimizer with a learning rate of $10^{-3}$, momentum term $\beta_1 = 0.9$, $\epsilon = 10^{-8}$, weight decay $10^{-5}$, gradient clipping at $10.0$, and a warm-up schedule of $1000$ steps.
	
	The RVNP ELBO objective was trained with a batch size of $100$ for $N_{\rm obs}=10000$ for $1000$ epochs using the Adam optimizer with a learning rate of $10^{-3}$, momentum term $\beta_1 = 0.9$, $\epsilon = 10^{-8}$, weight decay $10^{-5}$, gradient clipping at $10.0$, and a warm-up schedule of $1000$ steps. Early stopping is applied with a patience of $50$ epochs. In the forward modeling of the posterior and the simulator, $K_{\text{obs samples}} = 4$ is the number of samples used in the importance weighting, and $8$ samples are used from the simulator to evaluate the MC estimate of the likelihood. 
	
	The fine-tuning step runs across all simulations for $100$ epochs with a batch size of $1024$ using the Adam optimizer with a learning rate of $10^{-3}$, momentum term $\beta_1 = 0.9$, $\epsilon = 10^{-8}$, weight decay $10^{-5}$, gradient clipping at $10.0$, and a warm-up schedule of $1000$ steps.
	
	$10\%$ of the data is reserved for validation in all parts of the pipeline. 
	
	Experiments ran on Google Colab using a single NVIDIA L4 GPU in a single-process setup (no distributed workers).

	\subsection{Evaluation Procedure}
	We use $N_{\rm obs}=1000$ unseen samples for the evaluation, and we use $2500$ posterior samples for the evaluation of the different metrics.
	\subsection{Posterior and simulator density architectures}
	In every task and using all algorithms, we use a masked autoregressive affine flow \citep{papamakarios2018maskedautoregressiveflowdensity} for the approximate posterior distribution $q_{\boldsymbol{\phi}}(\boldsymbol{ \theta}|\,\boldsymbol{ x})$ with $5$ flow layers each parameterized by a neural network with a single hidden layer of width $52$. In RVNP and NPE-SC we use a rational quadratic spline flow for the likelihood surrogate, with $5$ flow layers, where each spline is modeled between $[-10,10]$ using $15$ knots that are parameterized by a neural network with a single hidden layer of width $52$. The normalizing flows are implemented using the \texttt{flowjax} Python package. 
	\subsection{Training Time}\label{app:training_time}
	Table~\ref{tab:training_times} reports total end-to-end wall-clock training time for each algorithm and task under the training setup described in this appendix. Runs were executed on Google Colab with a single NVIDIA L4 GPU in a single-process setup (no distributed workers). The Colab runtime capacities were: System RAM $53.0$ GB, GPU RAM $22.5$ GB, and disk $112.6$ GB. In practice, weaker hardware can also be used, at the cost of longer training times.
	
	\begin{table}[h]
		\centering
		\caption{\textbf{Training time for each algorithm and task.}}
		\label{tab:training_times}
		\begin{tabular}{lcccc}
			\toprule
			Algorithm & CS & SIR & Pendulum & Spectra \\
			\midrule
			NPE & 377.31s & 401.91s & 356.98s & 378.75s \\
			NNPE & 467.96s & 143.64s & 543.64s & 454.19s \\
			NPE-MMD & 641.20s & 586.84s & 638.06s & 679.60s \\
			NPE-SC & 1485.50s & 957.85s & 1221.67s & 801.59s \\
			RVNP-G & 1607.29s & 1949.99s & 1752.91s & 1720.04s \\
			RVNP-L & 1375.54s & 1563.71s & 1370.50s & 1446.84s \\
			\bottomrule
		\end{tabular}
	\end{table}

	\subsection{Benchmark Algorithms}\label{app:benchmark}
	We compare RVNP against four benchmark posterior estimators in a likelihood-free setting. Let
	\[
	(\boldsymbol{\theta},\boldsymbol{x}) \sim p(\boldsymbol{\theta})\,p(\boldsymbol{x}\mid \boldsymbol{\theta}),
	\qquad
	\boldsymbol{x}_{\rm obs}\sim p^*(\boldsymbol{x}),
	\]
	where \(\boldsymbol{x}_{\rm obs}\) denotes observed data (possibly under misspecification).
	All benchmarks use the same conditional normalizing-flow family for the posterior, so differences arise from the training objective and robustness mechanism.
	
	\textbf{Neural Posterior Estimation (NPE).}
	NPE minimizes conditional negative log-likelihood:
	\begin{equation}
		\mathcal{L}_{\mathrm{NPE}}(\boldsymbol{\phi})
		=
		-\mathbb{E}_{(\boldsymbol{\theta},\boldsymbol{x})\sim p(\boldsymbol{\theta})p(\boldsymbol{x}\mid\boldsymbol{\theta})}
		\left[\log q_{\boldsymbol{\phi}}(\boldsymbol{\theta}\mid h_{\boldsymbol{\omega}}(\boldsymbol{x}))\right].
	\end{equation}
	This baseline has no explicit misspecification-aware regularization.
	
	\textbf{Noisy Neural Posterior Estimation (NNPE).}
	Following \citet{ward2022robust}, NNPE corrupts simulated data with a spike-and-slab noise model and trains NPE on corrupted inputs. For coordinate \(j\),
	\[
	\tilde{x}_{j}=x_{j}+\varepsilon_{j},
	\qquad
	\varepsilon_{j}=(1-b_{j})\,\varepsilon^{\mathrm{spike}}_{j}+b_{j}\,\varepsilon^{\mathrm{slab}}_{j},
	\qquad
	b_{j}\sim \mathrm{Bernoulli}(\pi).
	\]
	A standard spike-Gaussian / slab-heavy-tail specification is
	\[
	\varepsilon^{\mathrm{spike}}_{j}\sim \mathcal{N}(0,\sigma_{\mathrm{spike}}^{2}),
	\qquad
	\varepsilon^{\mathrm{slab}}_{j}\sim \mathrm{Cauchy}(0,\gamma_{\mathrm{slab}}),
	\]
	equivalently,
	\[
	p(\varepsilon_{j})
	=
	(1-\pi)\,\mathcal{N}(0,\sigma_{\mathrm{spike}}^{2})
	+
	\pi\,\mathrm{Cauchy}(0,\gamma_{\mathrm{slab}}).
	\]

	The NNPE objective is
	\begin{equation}
		\mathcal{L}_{\mathrm{NNPE}}(\boldsymbol{\phi})
		=
		-\mathbb{E}_{(\boldsymbol{\theta},\boldsymbol{x})}
		\mathbb{E}_{\tilde{\boldsymbol{x}}\sim p(\tilde{\boldsymbol{x}}\mid \boldsymbol{x})}
		\left[\log q_{\boldsymbol{\phi}}(\boldsymbol{\theta}\mid h_{\boldsymbol{\omega}}(\tilde{\boldsymbol{x}}))\right].
	\end{equation}
	
	\textbf{Self-consistency NPE (NPE-SC).}
	Following \citet{mishra2025robust}, NPE-SC augments NPE with a self-consistency variance penalty. Define
	\[
	\log w(\boldsymbol{\theta},\boldsymbol{x}_{\rm obs})
	=
	\log \hat{p}_{\boldsymbol{\eta}}(\boldsymbol{x}_{\rm obs}\mid \boldsymbol{\theta})
	+\log p(\boldsymbol{\theta})
	-\log q_{\boldsymbol{\phi}}(\boldsymbol{\theta}\mid h_{\boldsymbol{\omega}}(\boldsymbol{x}_{\rm obs})),
	\]
	where \(\hat{p}_{\boldsymbol{\eta}}(\boldsymbol{x}\mid\boldsymbol{\theta})\) is a surrogate likelihood model. Then
	\begin{equation}
		\mathcal{L}_{\mathrm{NPE\text{-}SC}}(\boldsymbol{\phi})
		=
		\mathcal{L}_{\mathrm{NPE}}(\boldsymbol{\phi})
		+
		\lambda_{\mathrm{sc}}\,
		\mathbb{E}_{\boldsymbol{x}_{\rm obs}\sim p^*(\boldsymbol{x})}
		\!\left[
		\mathrm{Var}_{\boldsymbol{\theta}\sim q_{\boldsymbol{\phi}}(\cdot\mid h_{\boldsymbol{\omega}}(\boldsymbol{x}_{\rm obs}))}
		\big(\log w(\boldsymbol{\theta},\boldsymbol{x}_{\rm obs})\big)
		\right].
	\end{equation}
	We follow the reference implementation in:\\
	\texttt{https://github.com/bayesflow-org/self-consistency-real}
	for model configuration and training choices, including the recommended linear schedule for \(\lambda_{\mathrm{sc}}\) during training. The main differences in our implementation are that (i) we pre-train the surrogate likelihood \(\hat{p}_{\boldsymbol{\eta}}(\boldsymbol{x}\mid\boldsymbol{\theta})\) (and reuse the same likelihood surrogate as RVNP for fair comparison), and (ii) we use a bounded-domain transformation for parameters rather than \(\boldsymbol{\theta}\)-clipping.
	
	\textbf{NPE-MMD.}
	Following \citet{elsemuller2025doesunsuperviseddomainadaptation}, NPE-MMD jointly learns posterior and representation by matching simulated and observed feature distributions:
	\begin{equation}
		\mathcal{L}_{\mathrm{NPE\text{-}MMD}}(\boldsymbol{\phi},\boldsymbol{\omega})
		=
		-\mathbb{E}_{(\boldsymbol{\theta},\boldsymbol{x})}
		\left[\log q_{\boldsymbol{\phi}}(\boldsymbol{\theta}\mid h_{\boldsymbol{\omega}}(\boldsymbol{x}))\right]
		+
		\lambda_{\mathrm{mmd}}\,
		\mathrm{MMD}^{2}\!\left(
		\{h_{\boldsymbol{\omega}}(\boldsymbol{x})\},
		\{h_{\boldsymbol{\omega}}(\boldsymbol{x}^{\star})\}
		\right).
	\end{equation}
	For kernel \(k\), the population form is
	\[
	\mathrm{MMD}^{2}(P,Q)
	=
	\mathbb{E}_{\boldsymbol{z},\boldsymbol{z}'\sim P}[k(\boldsymbol{z},\boldsymbol{z}')]
	+
	\mathbb{E}_{\bar{\boldsymbol{z}},\bar{\boldsymbol{z}}'\sim Q}[k(\bar{\boldsymbol{z}},\bar{\boldsymbol{z}}')]
	-
	2\mathbb{E}_{\boldsymbol{z}\sim P,\bar{\boldsymbol{z}}\sim Q}[k(\boldsymbol{z},\bar{\boldsymbol{z}})].
	\]
	We obtain good results with a sum of inverse multiquadric kernels as in \cite{elsemuller2025doesunsuperviseddomainadaptation}.
	
	For fair comparison, \(h_{\boldsymbol{\omega}}\) is fixed (pre-computed/frozen) for NPE, NNPE, and NPE-SC, while NPE-MMD optimizes \(h_{\boldsymbol{\omega}}\) jointly with \(q_{\boldsymbol{\phi}}\) because robustness is induced through representation alignment.
	\paragraph{Default benchmark hyperparameters (robustness weights).}
	In benchmark runs, the regularization weights are set to:
	\[
	\lambda_{\mathrm{sc}} = 1.0
	\quad\text{(NPE-SC)},\qquad
	\lambda_{\mathrm{mmd}} = 1.0
	\quad\text{(NPE-MMD)}.
	\]
	NPE has no additional robustness weight beyond the NPE likelihood term.  
	NNPE likewise has no explicit \(\lambda\)-regularizer; robustness is controlled by the fixed corruption model parameters (default benchmark setting: \(\sigma_{\mathrm{spike}}=0.01\), slab scale \(=1.0\), and Bernoulli mixture probability \(\pi=0.5\)).

	\section{Evaluation Metrics}\label{app:metrics}
	
	In our evaluation metrics we consider quantities that capture both posterior robustness (coverage/calibration) and predictive discrepancy across the evaluated tasks. Let
	\[
	T=\{(\boldsymbol{\theta}^{*(i)},\boldsymbol{x}_{\rm obs}^{(i)})\}_{i=1}^{N_{\rm test}}
	\]
	denote a labeled test set.
	Unless otherwise stated, reported error bars are mean $\pm$ standard deviation computed across the full test set \(T\).
	
	\textbf{(i) Log posterior density (LPD).}
	For each test pair we evaluate
	\[
	\mathrm{LPD}^{(i)}=\log q_{\boldsymbol{\phi}}\!\left(\boldsymbol{\theta}^{*(i)}\mid \boldsymbol{x}_{\rm obs}^{(i)}\right),
	\]
	and report its mean and standard deviation across \(T\).
	
	\textbf{(ii) Average expected posterior coverage (AEPC).}
	Following \citet{hyndman1996hdr}, we evaluate calibration through an HDR coverage curve and a signed scalar summary. For each test case \(i\), let
	\[
	s^{(i)}(\boldsymbol{\theta})=\log q_{\boldsymbol{\phi}}(\boldsymbol{\theta}\mid \boldsymbol{x}_{\rm obs}^{(i)})
	\]
	denote the posterior score, and let \(s^{*(i)}=s^{(i)}(\boldsymbol{\theta}^{*(i)})\). Using posterior samples, we sort scores in descending order and define the HDR threshold at level \(\ell\in[0,1]\) by the corresponding score quantile. This induces the indicator
	\[
	c^{(i)}(\ell)=\mathbf{1}\!\left\{s^{*(i)} \ge \tau^{(i)}(\ell)\right\},
	\]
	i.e. whether the true parameter lies inside the \(\ell\)-level HDR.
	
	Aggregating over the test set gives the empirical expected coverage curve
	\[
	\widehat{\mathrm{Cov}}_{\rm HDR}(\ell)
	=\frac{1}{N_{\rm test}}\sum_{i=1}^{N_{\rm test}} c^{(i)}(\ell).
	\]
	Perfect calibration corresponds to \(\widehat{\mathrm{Cov}}_{\rm HDR}(\ell)=\ell\) for all \(\ell\). We summarize deviation from this diagonal by
	\[
	\delta_{\rm AEPC}
	=
	\int_0^1
	\left(\widehat{\mathrm{Cov}}_{\rm HDR}(\ell)-\ell\right)
	\, d\ell,
	\]
	and report this quantity as AEPC. Thus, \(\mathrm{AEPC}=0\) indicates ideal calibration, \(\mathrm{AEPC}<0\) indicates overconfidence, and \(\mathrm{AEPC}>0\) indicates underconfidence.
	
	\textbf{(iii) Standardized root mean squared error (SRMSE).}
	SRMSE is computed from posterior samples in parameter space after standardizing each dimension by its prior scale:
	\[
	\mathrm{SRMSE}^{(i)}
	=\frac{1}{N_s}\sum_{s=1}^{N_s}
	\sqrt{\frac{1}{d_\theta}\sum_{j=1}^{d_\theta}
		\left(
		\frac{\theta^{(i)}_{s,j}-\theta^{*(i)}_j}{\sigma_{\mathrm{prior},j}}
		\right)^2 }.
	\]
	
	\textbf{(iv) Posterior predictive distance (PPD).}
	We evaluate predictive discrepancy in observed-data space using a Euclidean distance between the observed point and posterior predictive samples. For each test case \(i\), draw posterior samples \(\boldsymbol{\theta}^{(i)}_s\sim q_{\boldsymbol{\phi}}(\boldsymbol{\theta}\mid \boldsymbol{x}_{\rm obs}^{(i)})\), then simulator predictive samples \(\tilde{\boldsymbol{x}}^{(i)}_s\sim p_{\boldsymbol{\Psi}}(\boldsymbol{x}\mid \boldsymbol{\theta}^{(i)}_s)\). We define
	\[
	\mathrm{PPD}^{(i)}
	=
	\frac{1}{N_s}\sum_{s=1}^{N_s}
	\left\lVert
	\boldsymbol{x}_{\rm obs}^{(i)}-\tilde{\boldsymbol{x}}^{(i)}_s
	\right\rVert_2.
	\]
	Smaller values indicate better posterior predictive fit.
	
	\textbf{(v) Posterior contraction (PC).}
	Posterior contraction measures how much uncertainty shrinks from prior to posterior in parameter space. For each test case \(i\), we define
	\[
	\mathrm{PC}^{(i)}
	=
	1-\frac{1}{d_\theta}\sum_{j=1}^{d_\theta}
	\min\!\left(1,\frac{\sigma^{(i)}_{\mathrm{post},j}}{\sigma_{\mathrm{prior},j}}\right),
	\]
	where \(\sigma^{(i)}_{\mathrm{post},j}\) is the posterior standard deviation of parameter \(j\) for case \(i\), and \(\sigma_{\mathrm{prior},j}\) is the prior standard deviation for that parameter. Larger values indicate stronger contraction (greater information gain from data), while values near zero indicate little update from the prior.

	\section{Experiment Details}\label{app:tasks}\label{app:exp_details}
	In this section we give a detailed description of the experiments carried out in the paper and the benchmark methods RVNP was evaluated against.

	\subsection{Task A: CS (Cancer and Stromal Cells)}\label{app:cs_task}
	The cancer and stromal cell development task from \citet{ward2022robust} models cell development in 2D space based on cell locality relative to unobserved parents. The simulator is conditional on three Poisson rate parameters $\boldsymbol{\theta}=(\lambda_c,\lambda_p,\lambda_d)$ with priors $\lambda_c\sim U(200,1500)$, $\lambda_p\sim U(3,20)$, and $\lambda_d\sim U(10,20)$. The total number of cells $N^c$, number of unobserved parents $N^p$, and the number of daughter cells for each parent $N_i^d$ are sampled as $N^c\sim \rm{Poisson}(\lambda_c)$, $N^p\sim \rm{Poisson}(\lambda_p)$, and $N^d_i\sim \rm{Poisson}(\lambda_d)$ for $i=1,...,N^p$. Cell locations $\{c_i\}_{i=1}^{N_c}$ and disease origin points $\{p_i\}_{i=1}^{N_p}$ are sampled uniformly across the 2D domain using homogeneous spatial point processes. For each origin point $p_i$, $r_i$ is the Euclidean distance to its $N_d^i$-th nearest cell. Cells falling within or on the boundary of this radius from $p_i$ are designated as cancerous. Distance-based summary measures are estimated by randomly sampling 50 stromal cells. The summary statistics are: N Cancer and N Stromal (number of cancer and stromal cells), Mean Min Dist and Max Min Dist (mean and maximum distance from stromal cells to their nearest cancer cell). 
	
	\textit{Misspecification:} Observed data have cells removed in tumor core regions (mimicking necrosis).

	\subsection{Task B: SIR (Susceptible-Infected-Recovered)}\label{app:sir_task}
	The stochastic epidemic model from \cite{ward2022robust} infers $\boldsymbol{\theta}=(\beta,\gamma)$, the time-varying infection rate and recovery rate. We draw $\beta\sim U(0,0.5)$ and $\gamma\sim U(0,0.5)$ independently, rejecting draws with $\gamma>\beta$ because these frequently yield trajectories with almost no infections. The retained prior has support $0<\gamma\leq\beta<0.5$. The SIR model emulates disease transmission dynamics: $
	\dot s = -\beta si$, $ 
	\dot i= \beta si - \gamma i$, and $ 
	\dot r = \gamma i. $
	A stochastic extension uses time-dependent transmission via variable infection rate $\tilde{\beta}_t$, characterized by the basic reproduction number $R_{0t} = \frac{\tilde{\beta}_t}{\gamma}$, which follows the mean-reverting SDE:
	$dR_{0t} = \eta\left(\beta \gamma^{-1} - R_{0t}\right)dt + \sigma R_0 dW_t$,
	where $\eta = 0.05$ and $\sigma = 0.05$ are fixed. The summary statistics are: mean, median and maximum infections, day of maximum infections, day of half total infections, and mean autocorrelation (lag 1). To make the summary statistics less hand-crafted, we apply an invertible linear transformation to the summary statistics as $\boldsymbol{ x}\mapsto \boldsymbol{Ax}$ where $\boldsymbol A$ has nonzero determinant. This transformation preserves the sufficiency of the statistic.

	\textit{Misspecification:} Weekend reporting delays reduce weekend counts by 5\%, added to Monday.

	\subsection{Task C: Pendulum}\label{app:pendulum_task}
	
	This task infers the fundamental frequency and amplitude of an oscillatory trajectory from noisy time-series observations. For each simulation, parameters are sampled as
	\[
	\omega_0\sim U(0.5,3.0),\qquad A\sim U(3.5,10.0),\qquad \phi\sim U(-0.03,0.03),
	\]
	where $\phi$ is a latent phase offset used to induce mild stochastic variation across trajectories. We then generate a clean horizontal-position signal
	\[
	f(t)=A\cos(\omega_0 t+\phi)
	\]
	on a fixed grid of 200 time steps over $10\,\mathrm{s}$, giving
	\[
	\boldsymbol{x}=(f(t_1),\ldots,f(t_{200}))\in\mathbb{R}^{200}.
	\]
	As in the main experiments, inference is performed on pre-trained neural embedding statistics rather than raw time series (Appendix \ref{app:nse}).
	
	\textit{Misspecification:} The simulator and observation processes differ only in observation noise scale. Simulator samples are corrupted with additive Gaussian noise at $5\%$, while observed data are generated with additive Gaussian noise at $20\%$. This creates a controlled signal-to-noise misspecification in which the simulator underestimates observational uncertainty.

	\subsection{Task D: Spectrum}\label{app:spectra_task}
	We study medium-resolution stellar spectra \citep{high_gal_me} where observations can fall in low-prior-probability regions that are poorly represented by finite simulation budgets. Synthetically generating observables in stellar astronomy given physical properties requires two simulation models: a stellar evolution and atmosphere model \citep{high_gal_me} which uses physical modeling to generate realistic stellar parameters of stars, and a synthetic atmosphere model which generates stellar spectra given atmospheric parameters. We use the MIST stellar evolution models \citep{MIST} to generate stellar parameters for stellar spectra inference and medium-resolution synthetic spectral templates \citep{2003IAUS.210P.A20C}, to generate a simulation pipeline mapping $\boldsymbol{ \theta}=(T_{\rm eff},\log g,\rm{[Fe/H]})$ --- the effective temperature, log-surface gravity, and metallicity --- to the synthetic spectrum $\boldsymbol{ x}$. In implementation, we use the \texttt{pystellibs} Python package (\url{https://mfouesneau.github.io/pystellibs/}) together with the \texttt{minimint} Python package (\url{https://github.com/segasai/minimint}) to generate the synthetic data. We assume a standard Milky Way prior on $\boldsymbol{ \theta}$ (see \cite{o2026data} for a discussion). We pre-train neural embeddings as outlined in Appendix \ref{app:nse}.

			\textit{Misspecification:} We generate observations from a metal-poor stellar stream \citep{martin2022stellar}. The stream's parameter distribution differs from the Galactic prior used to generate training simulations, constituting prior misspecification while the forward model remains unchanged. These observations also lie in low-probability regions under the simulation prior, so the finite simulation set contains few comparable samples. This creates an accompanying finite-simulation coverage gap between simulated training data and observed spectra. Although prior misspecification is not the main target of RVNP, this supplementary experiment shows that RVNP is robust under this prior shift.

	\subsection{Coverage Plots}
	We display the expected posterior coverage versus the empirical credible level from the benchmark tasks to give a more illustrative overview of the coverage.
		\begin{figure}[h]
		\centering
		\includegraphics[width=\linewidth]{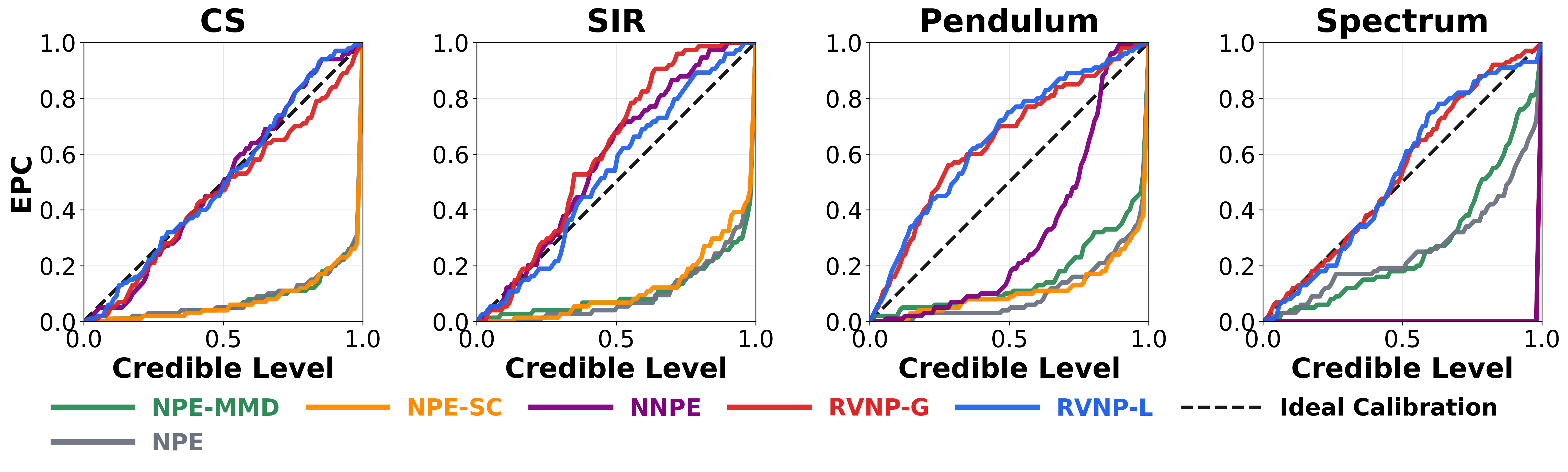}
			\caption{\textbf{Coverage curves across tasks.}}
		\label{fig:coverage}
	\end{figure}

	\section{Well-Specified SIR task}\label{app:wellspecified_sir}
	In this section, we evaluate the benchmark algorithms on the well-specified version of the SIR task. That is, observations are i.i.d.\ draws from the simulator sampling distribution, so there is no misspecification by construction.
	Table \ref{tab:sir-wellspec-eval} summarizes calibration and accuracy in this regime. As expected, standard NPE performs well on LPD and SRMSE when the simulator is exactly correct. RVNP remains competitive while retaining the same error-modeling mechanism used in the misspecified benchmarks, providing a useful sanity check that the learned correction does not lead to pathological behavior when no correction is needed.
	
\begin{table}[h]
	\centering
	\small
	\caption{\textbf{SIR well-specified evaluation summary.} }
	\begin{tabular}{lcccc}
		\hline
		Method & AEPC & LPD & PPD & SRMSE \\
		\hline
		RVNP-G & 0.213 & 1.406 $\pm$ 1.111 & 0.992 $\pm$ 0.533 & 0.063 \\
		RVNP-L & 0.129 & 2.018 $\pm$ 1.543 & 0.936 $\pm$ 0.598 & 0.055 \\
		NPE & 0.023 & 3.185 $\pm$ 2.128 & 0.869 $\pm$ 0.595 & 0.047 \\
		NNPE & 0.221 & 1.428 $\pm$ 0.983 & 191.550 $\pm$ 1562.438 & 0.072 \\
		NPE-MMD & 0.039 & 3.176 $\pm$ 2.105 & 0.922 $\pm$ 0.7 & 0.60 \\
		NPE-SC & -0.027 & 3.314 $\pm$ 2.333 & 0.839 $\pm$ 0.584 & 0.043 \\
		\hline
	\end{tabular}
	\label{tab:sir-wellspec-eval}
\end{table}

	\section{Additional Ablations and Sensitivity Analyses}\label{app:ablations}
	This appendix reports additional experiments that test sensitivity of RVNP to modeling and optimization choices. Unless otherwise stated, each ablation holds fixed the simulator, dataset, simulation/observation budgets, architectures, and optimization hyperparameters described in Appendix \ref{app:training_steps}, varying only the factor under study. We evaluate each ablation on the CS benchmark task. Overall, RVNP becomes empirically robust on this benchmark once $N_{\rm obs}\gtrsim 10$; ENPE fine-tuning provides a small additional robustness gain; and the method is broadly stable across the tested importance-sampling and inner Monte Carlo settings.

	\subsection{Robustness versus number of observations (CS)}\label{app:full_results}\label{app:ablation_nobs}
\begin{figure}[h]
		\centering
		\includegraphics[width=\linewidth]{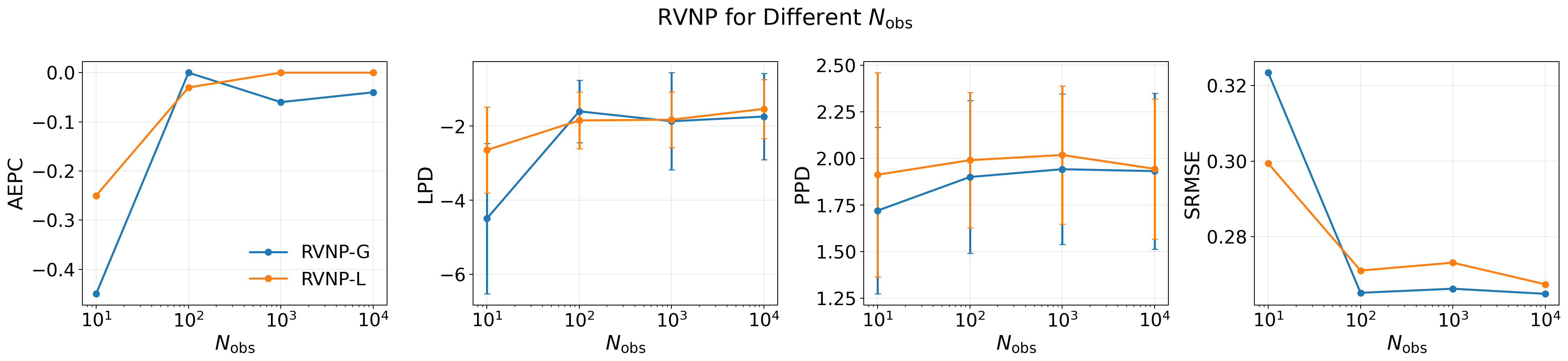}
		\caption{\textbf{Robustness versus number of observations.} RVNP becomes consistently robust on this benchmark once $N_{\rm obs}\gtrsim 10$, showing that a modest set of unlabeled observations is sufficient to constrain the learned correction.}
		\label{fig:nobs}
		\end{figure}

	In this section, we evaluate the effectiveness of RVNP when there are different numbers of observations to amortize over. This experiment quantifies how many unlabeled observations are needed for the error model to reliably fit the simulation-to-reality gap in the CS benchmark. Figure~\ref{fig:nobs} shows the robustness trend versus $N_{\rm obs}$.
	We observe a clear transition around $N_{\rm obs}\approx 10$: beyond this point, both RVNP variants maintain near-zero AEPC while retaining strong LPD/SRMSE and contraction. This supports the main motivation of RVNP: unlabeled observations provide a usable signal to constrain the simulation-to-reality gap through the learned error model.

	\subsection{Stability across training seeds (CS)}\label{app:ablation_seeds}
	Table~\ref{tab:cs_training_seeds} reports the CS results for five training seeds with the remaining experimental setup held fixed. Both RVNP variants show similar calibration and accuracy across these runs.
	\begin{table}[h]
		\centering
		\small
		\caption{\textbf{CS results across training seeds.} Each row reports one training run. AEPC is best near zero; higher LPD and lower PPD and SRMSE are better.}
		\label{tab:cs_training_seeds}
		\begin{tabular}{lrrrrrr}
			\toprule
			Method & Seed & AEPC & LPD & PPD & SRMSE & PC \\
			\midrule
			RVNP-L & 42 & $-0.00$ & $-1.53$ & $1.72$ & $1.76$ & $0.48$ \\
			RVNP-L & 43 & $-0.02$ & $-1.51$ & $1.70$ & $1.78$ & $0.47$ \\
			RVNP-L & 44 & $0.05$  & $-1.53$ & $1.80$ & $1.75$ & $0.47$ \\
			RVNP-L & 45 & $-0.01$ & $-1.49$ & $1.72$ & $1.76$ & $0.48$ \\
			RVNP-L & 46 & $-0.00$ & $-1.51$ & $1.71$ & $1.76$ & $0.48$ \\
			\midrule
			RVNP-G & 42 & $-0.04$ & $-1.74$ & $1.71$ & $1.75$ & $0.49$ \\
			RVNP-G & 43 & $-0.07$ & $-1.69$ & $1.72$ & $1.75$ & $0.48$ \\
			RVNP-G & 44 & $0.05$  & $-1.70$ & $1.81$ & $1.76$ & $0.48$ \\
			RVNP-G & 45 & $0.01$  & $-1.78$ & $1.79$ & $1.79$ & $0.49$ \\
			RVNP-G & 46 & $0.00$  & $-1.74$ & $1.71$ & $1.72$ & $0.48$ \\
			\bottomrule
		\end{tabular}
	\end{table}

	\subsection{Effect of likelihood-surrogate quality}\label{app:ablation_surrogate_quality}
\begin{figure}[h]
		\centering
		\includegraphics[width=\linewidth]{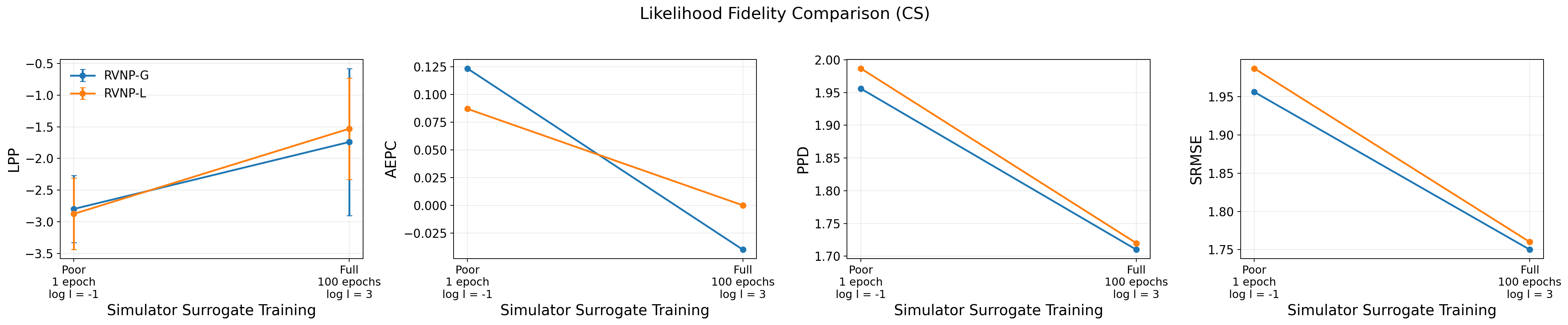}
		\caption{\textbf{Effect of likelihood-surrogate quality.} RVNP is stable once the learned surrogate reaches adequate fidelity, while poor surrogate fits degrade AEPC/LPD/SRMSE/PPD by conflating surrogate error with simulator-observation misspecification.}
		\label{fig:likelihood_surrogate}
	\end{figure}

	RVNP depends on the learned simulator likelihood surrogate $p_{\boldsymbol{\Psi}}(\boldsymbol{x}\mid\boldsymbol{\theta})$. In our experiments, we verified surrogate training via held-out simulated pairs and found RVNP performance stable once the surrogate reached a reasonable fit. This is the main modeling requirement of the implementation: if the surrogate is inaccurate, the learned correction can absorb surrogate error as well as genuine simulator-observation misspecification. Figure~\ref{fig:likelihood_surrogate} shows that AEPC/LPD/SRMSE/PPD degrade when surrogate fidelity is low.
	
	\subsection{Number of importance samples \(K\)}\label{app:ablation_importance_samples}
\begin{figure}[h]
		\centering
		\includegraphics[width=\linewidth]{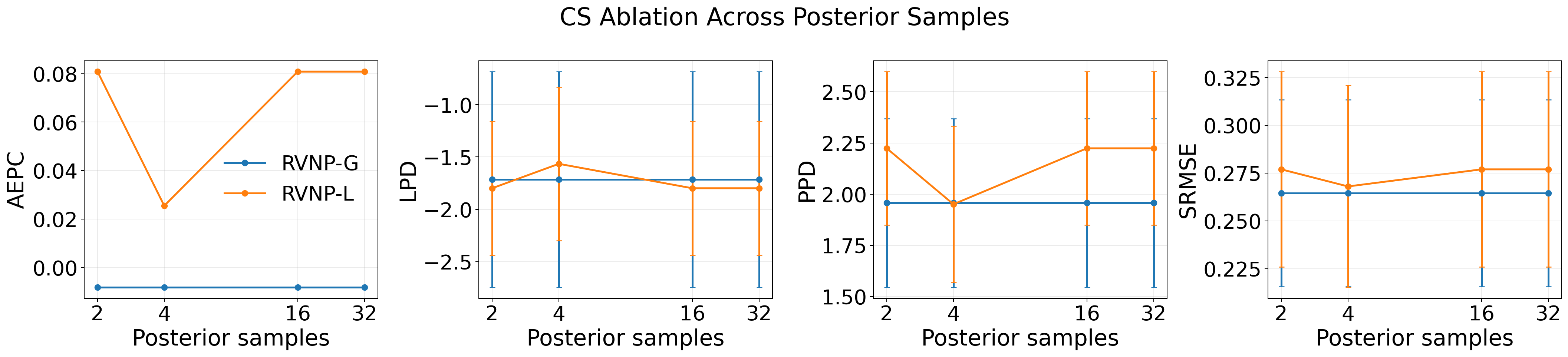}
			\caption{\textbf{Sensitivity to the number of importance samples \(K\).} Performance changes only mildly with \(K\); beyond moderate values the effect on AEPC/LPD/SRMSE/PPD is small.}
		\label{fig:importancesamples}
	\end{figure}

	We sweep the number of importance samples \(K\) used in the RVNP objective, holding the remaining training setup fixed. Figure \ref{fig:importancesamples} shows that the results are largely insensitive to \(K\); however, a higher $K$ does improve the stability of training.

	\subsection{Number of simulator Monte Carlo samples \(M\)}\label{app:ablation_inner_mc}
\begin{figure}[h]
		\centering
		\includegraphics[width=\linewidth]{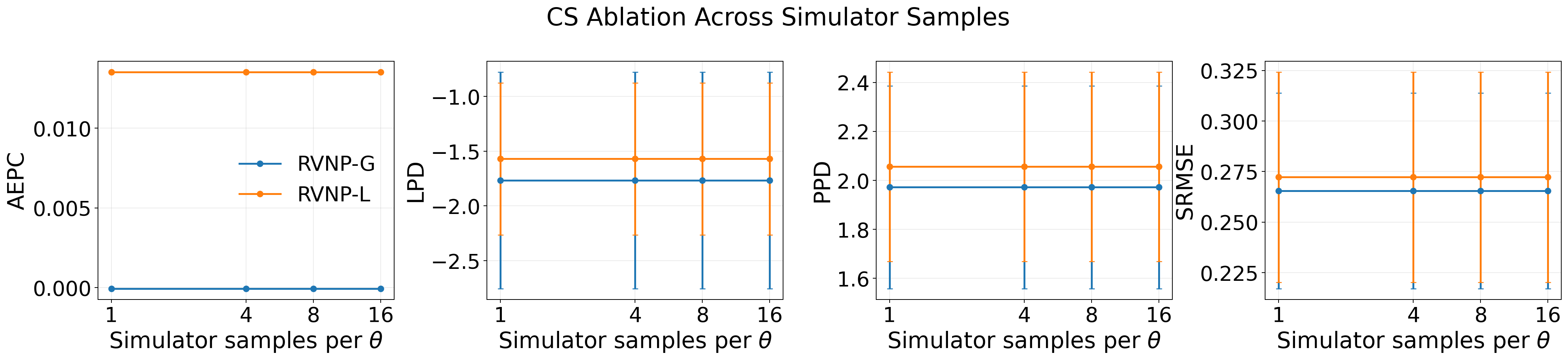}
			\caption{\textbf{Sensitivity to the number of inner Monte Carlo samples \(M\).} Performance is broadly stable across the tested range of \(M\), suggesting limited sensitivity to the variance of the inner expectation estimator.}
		\label{fig:inner_mc}
	\end{figure}

	We evaluate sensitivity to the number of inner Monte Carlo samples \(M\) used to estimate
	$\widehat{p}_{\boldsymbol{\alpha}}(\boldsymbol{x}_{\rm obs}\mid\boldsymbol{\theta})
	\approx
	\mathbb{E}_{p_{\boldsymbol{\Psi}}(\boldsymbol{x}\mid\boldsymbol{\theta})}[p_{\boldsymbol{\alpha}}(\boldsymbol{x}_{\rm obs}\mid\boldsymbol{x},\boldsymbol{\theta})]$
	inside the RVNP objective (Algorithm \ref{alg:rvnp_train}). Figure~\ref{fig:inner_mc} summarizes this sweep.
	Across the sweep, the downstream metrics change only weakly, indicating that the default inner Monte Carlo budget is sufficient in this setting.
	
	\subsection{Role of ENPE fine-tuning}\label{app:ablation_enpe}
\begin{figure}[h]
		\centering
		\includegraphics[width=\linewidth]{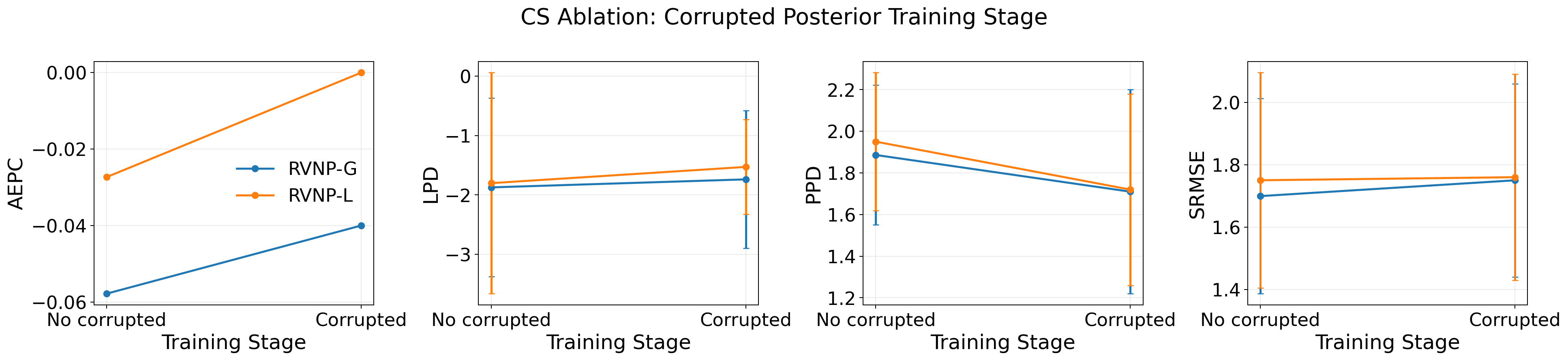}
			\caption{\textbf{Ablation of ENPE fine-tuning.} ENPE fine-tuning provides a small robustness gain, improving calibration and accuracy slightly relative to training without fine-tuning.}
		\label{fig:finetuning}
	\end{figure}

	We ablate the ENPE fine-tuning stage (Stage 5 in Appendix \ref{app:training_steps}) by comparing RVNP with and without fine-tuning, holding everything else fixed. Figure \ref{fig:finetuning} shows that ENPE provides a small but consistent improvement in robustness, improving calibration and accuracy slightly.
	\subsection{Shrinkage priors for the error model}\label{app:ablation_shrinkage}
\begin{figure}[h]
	\centering
	\includegraphics[width=\linewidth]{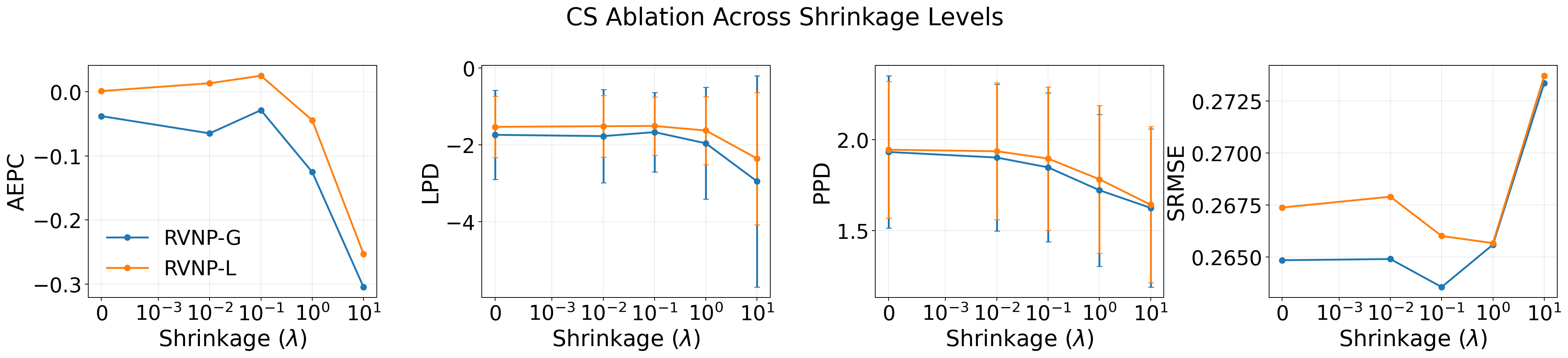}
	\caption{\textbf{Effect of Laplace shrinkage on the error model.} Increasing shrinkage reduces robustness under misspecification, consistent with limiting the learned correction and shifting residual discrepancy back onto $\boldsymbol{\theta}$.}
	\label{fig:shrinkage}
\end{figure}

	We study shrinkage/regularization of the error-model parameters $\boldsymbol{\alpha}$ (and hence of $\boldsymbol{\Sigma}(\cdot;\boldsymbol{\alpha})$), with the goal of encouraging minimal correction in well-specified regimes. Concretely, we use a Laplace shrinkage prior on the unconstrained covariance parameters (equivalently, an $\ell_1$ penalty) and vary its strength. Figure~\ref{fig:shrinkage} reports the resulting trend.
\subsection{Failure mode on very misspecified problem}\label{failure_mode}
One fundamental failure occurs when the observed and simulator distributions are indistinguishable in the chosen representation despite conditional misspecification. The unlabeled observation marginal then contains no signal that can identify a correction, so robustness requires stronger information such as paired calibration data or additional scientific constraints.

We also evaluated an adverse Damped Pendulum task in which observations are generated from the clean signal $f_{\rm damped}(t)=A e^{-0.1t}\cos(\omega_0 t+\phi)$ while the inference simulator retains the undamped signal $f(t)=A\cos(\omega_0 t+\phi)$ of Appendix \ref{app:pendulum_task}. This changes the trajectory mean systematically over time and is not representable by RVNP’s centered covariance-only Gaussian correction:
\begin{table}[t]
\centering
\caption{Results for the damped pendulum task.}
\begin{tabular}{llccccc}
\toprule
Task & Method & AEPC & LPD & PPD & SRMSE & PC \\
\midrule
Damped Pendulum & RVNP-L  & $-0.47$ & $-248.48 \pm 703.42$ & $0.60 \pm 0.11$ & $0.22 \pm 0.11$ & $0.98$ \\
                 & RVNP-G  & $-0.47$ & $-15.13 \pm 28.59$ & $0.78 \pm 0.13$ & $0.21 \pm 0.11$ & $0.96$ \\
                 & NPE     & $-0.47$ & $-718.78 \pm 1306.67$ & $0.22 \pm 0.13$ & $0.22 \pm 0.09$ & $1.00$ \\
                 & NPE-MMD & $-0.47$ & $-469.52 \pm 671.72$ & $0.31$ & $0.18 \pm 0.06$ & $1.00$ \\
                 & NPE-SC  & $-0.47$ & $-5293.21 \pm 9651.27$ & $0.22 \pm 0.14$ & $0.22 \pm 0.09$ & $1.00$ \\
                 & NNPE    & $-0.42$ & $-4.78 \pm 4.55$ & $13.78 \pm 13.18$ & $0.32 \pm 0.14$ & $0.74$ \\
\bottomrule
\end{tabular}
\end{table}
The mechanism is informative. In the correlator/summary representation, damping appears primarily as a smaller oscillation amplitude: observations generated at the true amplitude resemble undamped simulations with a lower amplitude and need not appear OOD. The unlabeled marginal therefore cannot distinguish simulator misspecification from a genuine change in $\boldsymbol{\theta}$, leaving no signal from which RVNP can identify the correction. This is a regime where stronger information, such as the parameter-labeled calibration set used by Wehenkel et al., is required. The experiment gives a concrete boundary: covariance inflation can handle substantial observable discrepancy, but cannot recover parameter information when misspecification changes its observational meaning.
\section{PIWAE objective and gradient estimator}\label{app:piwae}
This appendix summarizes the partially importance-weighted autoencoder (PIWAE) objective and its gradients \citep{2018arXiv180204537R}, specialized to our RVNP setting. The motivation for PIWAE is that while IWAE bounds become tighter as the number of importance samples $K$ increases, the signal-to-noise ratio of the gradient estimator for encoder/variational parameters can degrade for large $K$. PIWAE averages multiple smaller-IWAE objectives for the posterior update while retaining the full $K$-sample IWAE objective for the error-model update.

\subsection{Weights and IWAE objective}
For a single observation $\boldsymbol{x}_{\rm obs}$, define an importance weight
\begin{equation}
	w(\boldsymbol{\theta})
	:=
	\frac{p(\boldsymbol{\theta})}{q_{\boldsymbol{\phi}}(\boldsymbol{\theta}\mid \boldsymbol{x}_{\rm obs})}
	\,
	\widehat{p}_{\boldsymbol{\alpha}}(\boldsymbol{x}_{\rm obs}\mid\boldsymbol{\theta}),
\end{equation}
where $\widehat{p}_{\boldsymbol{\alpha}}(\boldsymbol{x}_{\rm obs}\mid\boldsymbol{\theta})$ denotes the Monte Carlo estimate of
$\mathbb{E}_{p_{\boldsymbol{\Psi}}(\boldsymbol{x}\mid\boldsymbol{\theta})}[p_{\boldsymbol{\alpha}}(\boldsymbol{x}_{\rm obs}\mid\boldsymbol{x},\boldsymbol{\theta})]$
used in the RVNP objective (Section \ref{methods}). The IWAE contribution for $\boldsymbol{x}_{\rm obs}$ with $K$ samples is
\begin{equation}
	\mathcal{L}_{\mathrm{IWAE}}^{(K)}
	=
	\log\left(\frac{1}{K}\sum_{k=1}^{K} w_k\right),
	\qquad
	w_k=w(\boldsymbol{\theta}_k),\ \boldsymbol{\theta}_k\sim q_{\boldsymbol{\phi}}(\boldsymbol{\theta}\mid\boldsymbol{x}_{\rm obs}).
\end{equation}

\subsection{PIWAE objective (\(K=GL\))}
PIWAE partitions the $K$ samples into $G$ groups of size $L$ such that $K=GL$. Let $\{\boldsymbol{\theta}_{g,\ell}\}$ denote samples with weights $w_{g,\ell}=w(\boldsymbol{\theta}_{g,\ell})$. The grouped objective for the posterior parameters $\boldsymbol{\phi}$ is
\begin{equation}
	\mathcal{L}_{\mathrm{PIWAE}}^{(G,L)}
	=
	\frac{1}{G}\sum_{g=1}^{G}
	\log\left(\frac{1}{L}\sum_{\ell=1}^{L} w_{g,\ell}\right).
\end{equation}
When $G=1$ this reduces to IWAE with $K=L$. For fixed compute budget $K$, the choice of the split $(G,L)$ can affect optimization behavior, and can be viewed as an additional hyperparameter of the gradient estimator.
In all experiments in this paper we fix $G=2$ and set $L=K/2$ (so that $K=2L$); we do not ablate over $G$.

\subsection{Gradients}
Define within-group normalized weights
\begin{equation}
	\tilde{w}_{g,\ell}
	:=
	\frac{w_{g,\ell}}{\sum_{\ell'=1}^{L} w_{g,\ell'}}.
\end{equation}
Using $\nabla \log \sum_{\ell} w_\ell = \sum_{\ell} \frac{w_\ell}{\sum_{\ell'}w_{\ell'}} \nabla \log w_\ell$, the posterior gradient takes the form
\begin{equation}
	\nabla_{\boldsymbol{\phi}}\mathcal{L}_{\mathrm{PIWAE}}^{(G,L)}
	=
	\frac{1}{G}\sum_{g=1}^{G}
	\sum_{\ell=1}^{L}
	\tilde{w}_{g,\ell}\,\nabla_{\boldsymbol{\phi}}\log w_{g,\ell},
\end{equation}
including dependence through the sampled $\boldsymbol{\theta}_{g,\ell}$ via reparameterization. The error-model parameters instead use the full $K$-sample IWAE gradient:
\begin{equation}
\nabla_{\boldsymbol{\alpha}}\mathcal{L}_{\mathrm{IWAE}}^{(K)}
=\sum_{k=1}^{K}\frac{w_k}{\sum_{j=1}^{K}w_j}\,\nabla_{\boldsymbol{\alpha}}\log w_k.
\end{equation} In practice, we compute these gradients with the reparameterization trick and automatic differentiation. To avoid notational collision with the inner Monte Carlo sample count used to estimate $\widehat{p}_{\boldsymbol{\alpha}}(\boldsymbol{x}_{\rm obs}\mid\boldsymbol{\theta})$ (Algorithm \ref{alg:rvnp_train}), we reserve $G$ and $L$ for the PIWAE grouping and keep $K$ for the total number of importance samples.

	\section{Error model covariance}\label{app:interpretability}
	 The learned covariance structure gives a direct summary of how RVNP explains the simulator-observation discrepancy in the chosen representation. For the SIR task, both the global and local error models assign substantially larger covariance to the final summary coordinates, with strong positive covariance between those coordinates. This indicates that RVNP primarily uses the error model to absorb the components most affected by the misspecification, while keeping the remaining summary coordinates comparatively precise for posterior inference.
	
\begin{equation}
	\Sigma_G = \begin{bmatrix}
		0.04 & 0.02 & 0.03 & -0.00 & 0.07 & 0.06 \\
		0.02 & 0.03 & 0.01 & 0.02 & 0.01 & -0.03 \\
		0.03 & 0.01 & 0.04 & -0.02 & 0.07 & 0.10 \\
		-0.00 & 0.02 & -0.02 & 0.16 & -0.08 & -0.27 \\
		0.07 & 0.01 & 0.07 & -0.08 & 2.13 & 2.27 \\
		0.06 & -0.03 & 0.10 & -0.27 & 2.27 & 2.69
	\end{bmatrix}
\end{equation}

\begin{equation}
	\Sigma_L(\boldsymbol{0}) = \begin{bmatrix}
		0.01 & 0.00 & 0.01 & 0.00 & -0.00 & -0.00 \\
		0.00 & 0.04 & 0.00 & 0.01 & 0.00 & -0.01 \\
		0.01 & 0.00 & 0.01 & 0.00 & -0.02 & -0.02 \\
		0.00 & 0.01 & 0.00 & 0.01 & 0.01 & 0.01 \\
		-0.00 & 0.00 & -0.02 & 0.01 & 1.95 & 2.01 \\
		-0.00 & -0.01 & -0.02 & 0.01 & 2.01 & 2.08
	\end{bmatrix}
\end{equation}
	
	Because the error model is explicit, these matrices can be inspected directly rather than inferred indirectly from posterior behavior. This is useful diagnostically: large learned covariance directions identify where the simulator requires correction, while the posterior still conditions on the remaining informative directions through the simulator likelihood surrogate.



  \section{Neural Embedding Architecture and Loss Function}\label{app:nse}

For the Pendulum and Spectra tasks, we learn a low-dimensional representation
\(
S_\omega(\boldsymbol{x}) \in \mathbb{R}^{d_z}
\)
with a separately parameterized decoder
\(
D_\phi
\).
The embedding is trained prior to posterior learning using simulated pairs
\(
\{(\boldsymbol{\theta}_i,\boldsymbol{x}_i)\}_{i=1}^B
\)
and a batch of observed inputs
\(
\{\boldsymbol{x}_{\mathrm{obs}}^{(j)}\}_{j=1}^{B_{\mathrm{obs}}}
\).
Only loss terms with nonzero coefficients are included in the objective.
We optimize the pendulum embedding model with Adam at learning rate $10^{-3}$ for 200 embedding pretraining epochs.

\subsubsection{Pendulum Task}

For pendulum, we use
\[
d_x = 200,\qquad d_\theta = 2,\qquad d_z = 5.
\]

The embedding pretraining objective includes only posterior and likelihood terms:
\[
\mathcal{L}_{\mathrm{emb}}^{\mathrm{pendulum}}
=
\lambda_{\mathrm{post}}\,\mathcal{L}_{\mathrm{post}}
+
\lambda_{\mathrm{like}}\,\mathcal{L}_{\mathrm{like}},
\]
with
\[
\lambda_{\mathrm{post}}=1,\qquad
\lambda_{\mathrm{like}}=1.
\]
Here, \(\mathcal{L}_{\mathrm{post}}\) is the posterior-flow loss on
\(q_\psi(\boldsymbol{\theta}\mid S_\omega(\boldsymbol{x}))\), and
\(\mathcal{L}_{\mathrm{like}}\) is the simulator-likelihood loss on
\(p_\eta(S_\omega(\boldsymbol{x})\mid \boldsymbol{\theta})\).

The pendulum encoder is a residual multilayer perceptron. The input trajectory
\(
\boldsymbol{x}\in\mathbb{R}^{200}
\)
is first projected into a hidden representation, passed through a stack of residual MLP
blocks with GELU nonlinearities, and then linearly mapped to a
5-dimensional latent code. The decoder mirrors this structure: the latent variable is
projected to the same hidden width, processed by residual MLP blocks, and finally mapped
back to the original 200-dimensional trajectory space. In all pendulum experiments we use
hidden width \(128\) and \(3\) residual blocks.
\subsubsection{Spectra Task}
The distance-covariance term is
\[
\mathcal{L}_{\mathrm{DC}}(\boldsymbol{\theta},\boldsymbol{z})
=
-\frac{\langle H_{\boldsymbol{\theta}},H_{\boldsymbol{z}}\rangle_F}
{\sqrt{\langle H_{\boldsymbol{\theta}},H_{\boldsymbol{\theta}}\rangle_F
		\,\langle H_{\boldsymbol{z}},H_{\boldsymbol{z}}\rangle_F+\varepsilon}},
\]
where
\(
\boldsymbol{z}=S_\omega(\boldsymbol{x})
\),
and
\(
H_{\boldsymbol{\theta}}
\)
and
\(
H_{\boldsymbol{z}}
\)
denote double-centered pairwise-distance matrices. Minimizing this objective encourages the
learned representation to preserve parameter-relevant structure.
For the spectra experiments, we use
\[
d_x = 250,
\qquad
d_\theta = 3,
\qquad
d_z = 5.
\]
The active embedding objective is
\[
\mathcal{L}_{\mathrm{emb}}^{\mathrm{spectra}}
=
\lambda_{\mathrm{rec}}
\frac{1}{B}\sum_{i=1}^{B}
\left\|
\boldsymbol{x}_i-D_\phi(S_\omega(\boldsymbol{x}_i))
\right\|_2^2
+
\lambda_{\mathrm{DC}}
\mathcal{L}_{\mathrm{DC}}(\boldsymbol{\theta},S_\omega(\boldsymbol{x})),
\]
with
\[
\lambda_{\mathrm{rec}} = 0.1,
\qquad
\lambda_{\mathrm{DC}} = 1.
\]

As in the pendulum task, the spectra encoder is a residual multilayer perceptron. The input
spectrum
\(
\boldsymbol{x}\in\mathbb{R}^{250}
\)
is linearly projected to a hidden representation, processed by a stack of residual MLP
blocks with GELU activations, and mapped to a 5-dimensional latent code. The decoder has
the same residual MLP form in reverse, mapping the latent code back to the 250-dimensional
input space. For spectra we use hidden width \(128\) and \(3\) residual blocks.

In both tasks, embedding models are optimized with Adam using learning rate \(10^{-3}\).

\input{checklist.tex}
\newpage

\end{document}

%% file: math_commands.tex
\usepackage{amsmath,amsfonts,bm}

\def\eqref#1{equation~\ref{#1}}

\def\1{\bm{1}}

\DeclareMathAlphabet{\mathsfit}{\encodingdefault}{\sfdefault}{m}{sl}
\SetMathAlphabet{\mathsfit}{bold}{\encodingdefault}{\sfdefault}{bx}{n}



%% file: checklist.tex
\section*{NeurIPS Paper Checklist}

\begin{enumerate}

\item {\bf Claims}
    \item[] Question: Do the main claims made in the abstract and introduction accurately reflect the paper's contributions and scope?
    \item[] Answer: \answerYes{} 
    \item[] Justification: The paper claims the proposed method produces a simulation-based inference method that is robust to misspecification by fitting the misspecification gap. We show this empirically on benchmarks and the contributions are accurately reflected by the results.

\item {\bf Limitations}
    \item[] Question: Does the paper discuss the limitations of the work performed by the authors?
    \item[] Answer: \answerYes{} 
    \item[] Justification: In the discussion we discuss the limitations and scope of the method introduced in the paper and the experiments and addressed some of the limitations in more detail through the appendices.

\item {\bf Theory assumptions and proofs}
    \item[] Question: For each theoretical result, does the paper provide the full set of assumptions and a complete (and correct) proof?
    \item[] Answer: \answerYes{} 
    \item[] Justification: A single proposition with trivial proof and standard assumptions included in the main text.

    \item {\bf Experimental result reproducibility}
    \item[] Question: Does the paper fully disclose all the information needed to reproduce the main experimental results of the paper to the extent that it affects the main claims and/or conclusions of the paper (regardless of whether the code and data are provided or not)?
    \item[] Answer: \answerYes{} 
    \item[] Justification: We describe the method in full in the main text with explicit algorithm and further details in the appendices. We extensively discuss the training details, architecture and datasets used in the experiments in the appendices. We believe we have included all necessary details for reproducibility. We further include our code in the supplementary materials.

\item {\bf Open access to data and code}
    \item[] Question: Does the paper provide open access to the data and code, with sufficient instructions to faithfully reproduce the main experimental results, as described in supplemental material?
    \item[] Answer: \answerYes{} 
    \item[] Justification: The data and code is available in the extra materials, and the readme describes how to reproduce the results.

\item {\bf Experimental setting/details}
    \item[] Question: Does the paper specify all the training and test details (e.g., data splits, hyperparameters, how they were chosen, type of optimizer) necessary to understand the results?
    \item[] Answer: \answerYes{} 
    \item[] Justification: Extensive details given in the appendices about all choices made to carry out the experiments. Key information is in the main text and references are made in the main text to the appendices.

\item {\bf Experiment statistical significance}
    \item[] Question: Does the paper report error bars suitably and correctly defined or other appropriate information about the statistical significance of the experiments?
    \item[] Answer: \answerYes{} 
    \item[] Justification: Yes, described in the metrics section in main text and appendix. Method for deriving error bars explicitly explained.

\item {\bf Experiments compute resources}
    \item[] Question: For each experiment, does the paper provide sufficient information on the computer resources (type of compute workers, memory, time of execution) needed to reproduce the experiments?
    \item[] Answer: \answerYes{} 
    \item[] Justification: We explicitly outline our own requirements used to generate the experiments.
    
\item {\bf Code of ethics}
    \item[] Question: Does the research conducted in the paper conform, in every respect, with the NeurIPS Code of Ethics \url{https://neurips.cc/public/EthicsGuidelines}?
    \item[] Answer: \answerYes{} 
    \item[] Justification: There are no ethical concerns to report.

\item {\bf Broader impacts}
    \item[] Question: Does the paper discuss both potential positive societal impacts and negative societal impacts of the work performed?
    \item[] Answer: \answerNA{} 
    \item[] Justification: There are no negative impacts. Potential positive societal impacts include aiding scientific advancement.
    
\item {\bf Safeguards}
    \item[] Question: Does the paper describe safeguards that have been put in place for responsible release of data or models that have a high risk for misuse (e.g., pre-trained language models, image generators, or scraped datasets)?
    \item[] Answer: \answerNA{} 
    \item[] Justification: The paper poses no such risks.

\item {\bf Licenses for existing assets}
    \item[] Question: Are the creators or original owners of assets (e.g., code, data, models), used in the paper, properly credited and are the license and terms of use explicitly mentioned and properly respected?
    \item[] Answer: \answerYes{} 
    \item[] Justification: All citations properly accounted for.

\item {\bf New assets}
    \item[] Question: Are new assets introduced in the paper well documented and is the documentation provided alongside the assets?
    \item[] Answer: \answerYes{} 
    \item[] Justification: A new method is introduced with full implementation details in code and theory.

\item {\bf Crowdsourcing and research with human subjects}
    \item[] Question: For crowdsourcing experiments and research with human subjects, does the paper include the full text of instructions given to participants and screenshots, if applicable, as well as details about compensation (if any)? 
    \item[] Answer: \answerNA{} 
    \item[] Justification:  \answerNA{}

\item {\bf Institutional review board (IRB) approvals or equivalent for research with human subjects}
    \item[] Question: Does the paper describe potential risks incurred by study participants, whether such risks were disclosed to the subjects, and whether Institutional Review Board (IRB) approvals (or an equivalent approval/review based on the requirements of your country or institution) were obtained?
    \item[] Answer:  \answerNA{} 
    \item[] Justification:  \answerNA{}

\item {\bf Declaration of LLM usage}
    \item[] Question: Does the paper describe the usage of LLMs if it is an important, original, or non-standard component of the core methods in this research? Note that if the LLM is used only for writing, editing, or formatting purposes and does \emph{not} impact the core methodology, scientific rigor, or originality of the research, declaration is not required.
    \item[] Answer: \answerNA{}
    \item[] Justification:  \answerNA{}

\end{enumerate}